\documentclass{article}

\usepackage[main,final,nonatbib]{neurips_2026} 
\usepackage[numbers,sort]{natbib}

\usepackage[utf8]{inputenc}
\usepackage[T1]{fontenc}
\usepackage{hyperref}
\usepackage{url}
\usepackage{booktabs}
\usepackage{amsfonts}
\usepackage{amsmath}
\usepackage{amssymb}
\usepackage{nicefrac}
\usepackage{microtype}
\usepackage{xcolor}
\usepackage{graphicx}
\usepackage{algorithm}
\usepackage{algorithmic}
\usepackage{subcaption}
\usepackage{multirow}
\usepackage{xspace}
\usepackage{enumitem}
\usepackage{colortbl}
\usepackage{amsthm}

\usepackage{adjustbox}

\newtheorem{theorem}{Theorem}
\newtheorem{assumption}{Assumption}
\newtheorem{definition}{Definition}
\newtheorem{proposition}{Proposition}

\theoremstyle{definition}
\newtheorem{remark}{Remark}

\newcommand{\method}{KV-PRM\xspace}
\newcommand{\steer}{KV Steering\xspace}

\newcommand{\pplus}{P(+)}
\newcommand{\eg}{e.g.,\xspace}
\newcommand{\ie}{i.e.,\xspace}

\setlist[itemize]{noitemsep,leftmargin=*,topsep=0pt}
\setlist[enumerate]{noitemsep,leftmargin=*,topsep=0pt}

\newcommand{\rev}[1]{#1}

\title{\method: Efficient Process Reward Modeling via KV-Cache Transfer for Multi-Agent Test-Time Scaling}

\author{
  Peng Kuang$^{1}$,
  Haibo Jin$^{1}$,
  Xiaoyu Han$^{1}$,
  Yanli Wang$^{2}$,
  Xiaopeng Yuan$^{1}$,\\
  \textbf{Ye Yu$^{1}$,}
  \textbf{Kaidi Xu$^{3}$,}
  \textbf{Haohan Wang$^{1}$}\\
  $^1$University of Illinois Urbana-Champaign, 
  $^2$Imperial College London, \\
  $^3$City University of Hong Kong\\
  \texttt{pengk3@illinois.edu,}
  \texttt{kaidixu@cityu.edu.hk, haohanw@illinois.edu} 
}

\begin{document}

\maketitle

\begin{abstract}
Process Reward Models (PRMs) have been proven to be highly effective in guiding test-time scaling (TTS) methods, which significantly boost the capabilities of LLM-based multi-agent systems.
However, existing PRMs are \emph{text-based}: they re-encode the entire trajectory text from scratch. In long multi-agent rollouts, the scoring cost, growing quadratically with respect to sequence length $L$, creates a severe computational bottleneck, severely limiting PRMs' application in long-context scenarios.
To resolve this, we introduce \method, a highly efficient process reward model that eliminates the heavy text re-encoding by directly reading the KV cache produced naturally during the LLM's generation phase. By processing a single "verify token" against the pre-existing KV cache, \method reduces the scoring cost from $O(L^2)$ to $O(L)$. 
\rev{We formally prove that the KV cache contains greater information capacity than text, and is more efficient for downstream reward modeling.}
Empirically, across the MATH, GSM8K, and AIME benchmarks \rev{and three model families}, \method matches text-PRMs
under various TTS methods such as Beam Search, MCTS, and Weighted Voting, with up to a 5,000$\times$ reduction in scoring FLOPs, a 37$\times$ reduction in latency, and a 34$\times$ reduction in per-sequence memory footprint compared to text-based PRMs. Our code is available at \url{https://github.com/KrumperIsOK/KV-PRM}.
\end{abstract}

\section{Introduction}
\label{sec:intro}

Scaling test-time compute has recently emerged as a highly effective direction for improving the capabilities of Large Language Models (LLMs) on complex reasoning tasks~\citep{wei2022chain, snell2024scaling, deepseek2025r1}. Instead of relying on a single greedy decoding pass, advanced reasoning pipelines leverage multi-agent systems (MAS)~\citep{guo2024multiagent} and test-time search (TTS) algorithms~\cite{kuang2026optimal}, such as beam search and Monte Carlo Tree Search (MCTS)~\citep{feng2024tsllm}, to explore diverse solution trajectories. A critical component guiding this search is the Process Reward Model (PRM)~\citep{lightman2023lets, wang2024mathshepherd, zheng2025prmsurvey}. By evaluating the correctness of intermediate reasoning steps, PRMs guide the search algorithm to allocate compute toward more promising partial solutions.

However, scaling test-time search exposes a critical computational bottleneck: the cost of the PRM itself. Existing PRMs are fundamentally \emph{text-based}. For every candidate trajectory proposed by the generator, a text-PRM must encode the entire sequence of text tokens from scratch to output a reward. As self-attention scales quadratically with sequence length, this introduces an $O(L^2)$ computational cost per scoring call, where $L$ is the sequence length. In modern MAS workflows, reasoning trajectories routinely span thousands or tens of thousands of tokens. When searching over dozens of candidates at multiple agent handoffs, the redundant text re-encoding by the PRM becomes an even heavier burden, often matching the FLOPs required for the generation itself. Such an architectural bottleneck limits the application of PRMs in long-context reasoning scenarios. 

In this work, we identify a missed opportunity in current test-time scaling systems: the generation process of agents inherently computes a rich, high-dimensional representation of the trajectory, the Key-Value (KV) cache. As autoregressive generation requires caching historical states to predict the next token, a complete record of the model's intermediate representations across all layers and positions is produced naturally during the LLM's generation phase. 
In Section~\ref{sec:theory}, we further theoretically prove the representational advantage of KV cache and its efficient exploitation compared to text.
\rev{Specifically, we establish a theoretical framework demonstrating that the KV cache contains greater information capacity than text, and requires much less compute for downstream reward modeling.}
We formally show that the KV cache provides an $\Omega(d / \log|\mathcal{V}|)$ capacity advantage per position over text tokens, and prove that the approximation gap decays exponentially in readout depth $k$, with efficient $k\!=\!1$ capturing most of extractable reward information for practical architectures (Theorem~\ref{thm:diminishing}).
\rev{Re-encoding the decoded text from scratch discards this high-fidelity, continuous representation, forcing the verification model to rely solely on discrete text tokens that collapse the rich internal state.}

Building upon this insight, we propose \method, a highly efficient architecture for process reward modeling. Instead of processing text, \method scores trajectories via \emph{KV-cache transfer}. By appending a single ``verify token'' to the end of the sequence and processing it through a lightweight LoRA adapter~\citep{hu2022lora} conditioned on the base model's pre-existing KV cache, \method achieves orders-of-magnitude complexity reduction. The per-call scoring cost drops from $O(L^2)$ to $O(L)$, resulting in up to a $5{,}000\times$ reduction in scoring FLOPs for typical MAS trajectories.


Empirically, we evaluate \method across three model scales (Qwen3-0.6B, 4B, 8B) \rev{and two additional model families (Llama-3.2-3B and Ministral-3-3B)} on MATH, GSM8K, and AIME benchmarks in Section~\ref{sec:experiments}. We find that \method not only recovers the performance of text-based PRMs but frequently outperforms them, closing the theoretical approximation gap while accelerating wall-clock scoring latency by $15$--$37\times$, and reducing the per-sequence memory footprint by 34.2$\times$. Finally, as \method evaluates the continuous KV cache rather than discrete text, its reward signal is fully differentiable. We explore this property through a proof-of-concept technique called \steer (Section~\ref{sec:steering}), demonstrating that gradient-based optimization of the latent messages between agents is structurally possible and presents a promising direction for test-time scaling in the latent space.
In summary, our contributions are:
\begin{itemize}
    \item \textbf{Theoretical framework:} We formalize the representational advantage of the KV cache over text tokens for verification, and prove that the approximation gap decays exponentially in readout depth, formally justifying the efficiency of KV cache-based reward modeling. 
    \item \textbf{Algorithm (\method):} We introduce a novel PRM architecture that reuses the generation KV cache, reducing scoring complexity from $O(L^2)$ to $O(L)$ by addressing the architectural bottleneck of text-based PRMs.
    \item \textbf{Extensive empirical validation:} We demonstrate that \method matches or exceeds the accuracy of text-based PRMs across multiple model scales\rev{, model families,} and search algorithms\rev{, transfers to code generation, and extends to generators of different architectures,} while delivering $37\times$ wall-clock speedups and 34.2$\times$ reduction in per-sequence memory footprint.
\end{itemize}

\section{Preliminaries and Notations}
\label{sec:prelim}

\textbf{Multi-agent reasoning systems.}
A multi-agent system (MAS) decomposes a problem across $D$ specialized LLM agents $a_1, \ldots, a_D$~\citep{guo2024multiagent, hong2024metagpt}.
Each agent $a_j$ generates output $\mathbf{o}_j$ conditioned on preceding agents' outputs and the original question, using a decoder-only transformer $f_\theta$ with $N_L$ layers, $n_h$ attention heads per layer, and hidden dimension $d$.
Let $L_j$ denote the total sequence length after agent $a_j$ completes, and $L = L_D$ the final trajectory length.

\textbf{Process reward models and test-time search.}
A process reward model (PRM) predicts a score $s_j \in [0, 1]$ for the trajectory up to agent step $j$, indicating the likelihood of reaching a correct answer~\citep{lightman2023lets, wang2024mathshepherd, yazdani2025masprm}.
PRM scores guide test-time search (TTS), such as beam search, MCTS~\citep{kocsis2006uct}, or weighted majority voting~\citep{wang2023selfconsistency}, to select high-quality trajectories~\citep{snell2024scaling}.
With beam width $W$ and $D$ agent steps, the PRM is invoked at least $W \cdot D$ times.
Existing PRMs are \emph{text-based}: they perform a full forward pass over the trajectory text $\mathbf{x}_j$ of length $L_j$, incurring cost $F_{\text{forward}}(L_j)$ per call.

\textbf{KV cache in autoregressive generation.}
During autoregressive generation, the transformer avoids redundant recomputation by caching the key and value projections from all previous tokens at every layer.
After generating $L$ tokens, this \emph{KV cache} $(\mathbf{K}, \mathbf{V})$ with $\mathbf{K}, \mathbf{V} \in \mathbb{R}^{N_L \times L \times d}$ constitutes a complete record of the model's intermediate representations across all layers and positions.
Each new token is generated by attending to this cache at cost $O(d \cdot L)$ per token.
Crucially, the KV cache is a \emph{byproduct} of generation that exists whether or not it is used for any downstream purpose.

\textbf{The computational bottleneck.}
The dominant cost in a forward pass over a sequence of length $L$ is:
\begin{equation}
    F_{\text{forward}}(L) = N_L \cdot \big(\underbrace{c_{\text{attn}} \cdot d \cdot L^2}_{\text{self-attention}} + \underbrace{c_{\text{ffn}} \cdot d^2 \cdot L}_{\text{feed-forward}}\big)
    \label{eq:flops_forward}
\end{equation}
For the long trajectories in MAS ($L \gg d$), self-attention dominates, and each text-PRM scoring call costs $O(d \cdot L^2)$.
When the PRM and the LLM are at the same scale, PRM scoring roughly \emph{doubles} the total system compute.
This motivates our central question: \emph{can we reduce the per-call scoring cost from $O(d \cdot L^2)$ to $O(d \cdot L)$, a reduction of $L\times$, without a loss in scoring quality?}
We begin by analyzing the information in the KV cache to establish that this reduction is not only computationally feasible but theoretically well-grounded in Section~\ref{sec:theory}, before presenting our method in Section~\ref{sec:method}.

\section{Theoretical Understanding}
\label{sec:theory}


We begin by reviewing the representational advantage of the KV cache over text tokens in Section \ref{sec:repr_advantage}, which serves as the foundation for our main theoretical contributions: a verification error decomposition \rev{into a representation-dependent error floor and a readout approximation gap} (Theorem \ref{thm:readout}), and a diminishing-returns analysis proving that a single-token readout readily captures most of the extractable reward information (Theorem \ref{thm:diminishing}). Full proof in Appendix~\ref{app:diminishing}.
\subsection{Representational Advantage of KV Cache}
\label{sec:repr_advantage}


\rev{As a foundation for the results that follow, we first establish that the KV cache produced during generation carries richer information than decoded text.}

\begin{assumption}[Linear Representation Hypothesis \cite{park2023linear,zou2025latentcollaborationmultiagentsystems}]
\label{assum:lrh}
The hidden embeddings $h \in \mathbb{R}^d$ produced by the transformer are linear combinations $h = \sum_{i=1}^{d} c_i s_i$ of a linearly independent semantic basis $\{s_1, \ldots, s_d\} \subset \mathbb{R}^d$ with ternary coefficients $c_i \in \{0, \pm 1\}$, where $c_i = 0$ indicates absence of semantic $i$, and $c_i = \pm 1$ its positive or negative presence.
\end{assumption}

\begin{proposition}[Representational Advantage of KV Cache]
\label{thm:repr_advantage}
Let $f_\theta$ be an autoregressive transformer that, during generation, produces KV cache $\mathbf{H} = (\mathbf{K}, \mathbf{V})$ and text tokens $\mathbf{x} = (x_1, \ldots, x_L)$ via $\mathbf{x} = \mathrm{decode}(\mathbf{H})$. Let $Y$ denote any target variable of interest (\eg trajectory correctness). Then, under Assumption~\ref{assum:lrh}, if the information in the KV cache of a length-$L$ trajectory is to be expressed losslessly through text, the required number of tokens is at least
    \begin{equation}
        m' \;=\; \Omega\!\left(\frac{d \cdot L}{\log |\mathcal{V}|}\right),
        \label{eq:capacity}
    \end{equation}
    where $|\mathcal{V}|$ is the vocabulary size. Equivalently, the KV cache is $\Omega(d / \log|\mathcal{V}|)$ times more information-dense per position than text.
\end{proposition}


\subsection{Efficient Verification via KV-Cache Readout}
\label{sec:readout}

\rev{Proposition}~\ref{thm:repr_advantage} establishes that the KV cache is an information-rich representation for verification. We now formalize how this richness can be exploited efficiently.

\begin{definition}[Depth-$k$ Readout]
\label{def:readout}
A \emph{depth-$k$ readout} $\mathcal{R}_k: (\mathbf{K}, \mathbf{V}) \mapsto s \in [0,1]$ processes $k$ query vectors against the pre-existing KV cache of length $L$ through a parameterized attention function $f_\psi$. We denote by $\mathcal{F}_k$ the function class of all depth-$k$ readouts with bounded parameter capacity $\|\psi\| \leq B$. Text re-encoding corresponds to $k = L$; KV-cache readout uses $k \ll L$.
\end{definition}

\begin{definition}[Approximation Gap]
\label{def:approx_gap}
For a function class $\mathcal{F}_k$ operating on KV cache $\mathbf{H}$, the \emph{approximation gap} is:
\begin{equation}
    \mathrm{Gap}(\mathcal{F}_k,\, \mathbf{H}) \;=\; \inf_{f \in \mathcal{F}_k} \mathbb{E}[\ell(f(\mathbf{H}),\, Y)] \;-\; \inf_{f}\, \mathbb{E}[\ell(f(\mathbf{H}),\, Y)],
    \label{eq:approx_gap}
\end{equation}
where the second infimum is over all measurable functions. $\mathrm{Gap}(\mathcal{F}_k, \mathbf{H})$ is monotonically non-increasing in $k$.
\end{definition}

\begin{theorem}[Verification Error Decomposition]
\label{thm:readout}
For any depth-$k$ readout $\mathcal{R}_k \in \mathcal{F}_k$, the verification error decomposes as:
\begin{equation}
    \epsilon(\mathcal{R}_k) = \epsilon_{\mathrm{Bayes}}(\mathbf{H}) + \mathrm{Gap}(\mathcal{F}_k,\, \mathbf{H}),
    \label{eq:error_decomp}
\end{equation}
where $\epsilon_{\mathrm{Bayes}}(\mathbf{H}) \leq \epsilon_{\mathrm{Bayes}}(\mathbf{x})$ by Proposition~\ref{thm:repr_advantage}. That is, the Bayes-optimal predictor from the KV cache achieves equal or lower error than the Bayes-optimal predictor from text. The approximation gap $\mathrm{Gap}(\mathcal{F}_k, \mathbf{H})$ is controlled by the readout depth $k$ and is monotonically non-increasing in $k$.
\end{theorem}


\begin{remark}
\label{rem:readout_implications}
Theorem~\ref{thm:readout} decompose the verification error to a lower bound determined by the information richness of the input and a gap between the empirical and optimal verifier, where the former favors KV cache given its representational adventage proven in Proposition~\ref{thm:repr_advantage}.
\end{remark}

\begin{assumption}[Low-Rank Reward Structure]
\label{assum:low_rank_reward}
The reward $Y$ depends on the KV cache $\mathbf{H}$ through a linear projection $\Phi(\mathbf{H}) = W_\phi \, \mathrm{pool}(\mathbf{H}) \in \mathbb{R}^r$, where $r$ is the effective reward dimensionality and $W_\phi \in \mathbb{R}^{r \times d}$.
The singular values $\sigma_1 \geq \sigma_2 \geq \cdots \geq \sigma_r$ of the covariance of $\Phi(\mathbf{H})$ projected onto the reward-relevant subspace decay as $\sigma_j = O(e^{-\alpha j})$ for some spectral decay rate $\alpha > 0$ \cite{fu2026theoretical,liu2026spectralgeometrythoughtphase}.
\end{assumption}

\begin{theorem}[Diminishing Returns of Readout Depth]
\label{thm:diminishing}
Under Assumptions~\ref{assum:lrh} and~\ref{assum:low_rank_reward}, let $\mathcal{R}_k \in \mathcal{F}_k$ be a depth-$k$ readout over KV cache $\mathbf{H}$, and define the marginal information gain $\Delta I_k = I(\mathcal{R}_k(\mathbf{H});\, Y \mid \mathcal{R}_{k-1}(\mathbf{H}))$. Then:
\begin{enumerate}[label=(\alph*)]
    \item \textbf{Exponential decay.} The marginal gain decays exponentially:
    \begin{equation}
        \Delta I_k \;\leq\; C_0 \cdot e^{-\alpha \cdot n_h N_L \cdot (k-1)},
        \label{eq:diminishing}
    \end{equation}
    where $C_0 = \sum_{j=1}^{r} \frac{1}{2}\log(1 + \mathrm{SNR} \cdot \sigma_j^2)$ is the total extractable information, with $\mathrm{SNR} = \mathrm{Var}(Y)/\sigma_\epsilon^2$ the signal-to-noise ratio.
    \item \textbf{$k\!=\!1$ near-optimality.} The fraction of reward information captured at $k=1$ satisfies:
    \begin{equation}
        \frac{I(\mathcal{R}_1(\mathbf{H});\, Y)}{I(\mathbf{H};\, Y)} \;\geq\; 1 - \frac{e^{-\alpha \cdot n_h N_L}}{1 - e^{-\alpha \cdot n_h N_L}}.
        \label{eq:k1_optimality}
    \end{equation}
    \item \textbf{Cost--information tradeoff.} The information gained per additional FLOP decays as:
    \begin{equation}
        \frac{\Delta I_k}{F_{\mathrm{readout}}(k,L) - F_{\mathrm{readout}}(k\!-\!1,L)} = O\!\left(\frac{e^{-\alpha \cdot n_h N_L \cdot (k-1)}}{d \cdot L}\right).
        \label{eq:cost_info}
    \end{equation}
\end{enumerate}
\end{theorem}

\begin{remark}
\label{rem:diminishing_implications}
Theorem~\ref{thm:diminishing} shows that reward modeling on KV cache is cost-efficient, where using a single query token is the most cost-efficient operating point, readily capturing most of the information.
\end{remark}

\section{Method}
\label{sec:method}

Motivated by the theoretical analysis in Section~\ref{sec:theory}, we introduce \method, a process reward model that instantiates the KV-cache readout framework (Definition~\ref{def:readout}) at $k = 1$, achieving the maximum $\Omega(L)$ speedup by scoring trajectories through the KV cache already produced during generation.

\begin{figure}[t]
    \centering
    \includegraphics[width=\linewidth]{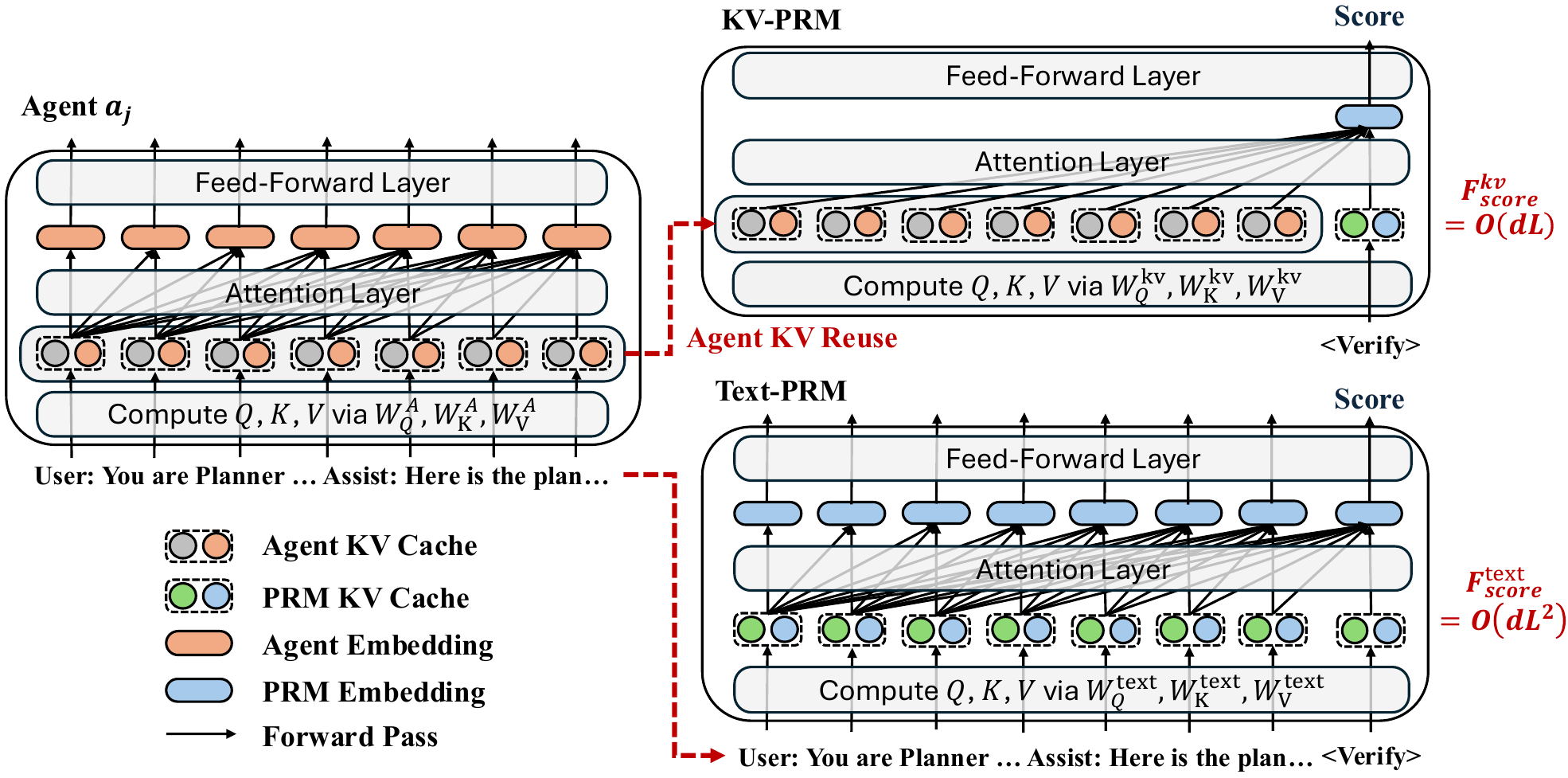}
    \caption{Text-PRM re-encodes the full trajectory for each scoring call at $O(dL^2)$ FLOPs. \method reuses the KV cache from generation and scores with a single verify token at $O(dL)$ FLOPs, a reduction of $L\times$, which is three orders of magnitude for typical MAS trajectories.}
    \label{fig:overview}
\end{figure}

\subsection{KV-PRM: Scoring via KV-Cache Transfer}
\label{sec:kvprm}

During generation, the LLM naturally produces a KV cache, encoding the full reasoning trajectory. \method reuses this pre-existing cache to score trajectory quality, without re-encoding.

\textbf{Architecture.}
Let $f_\theta$ denote the base language model with parameters $\theta$, and let $f_{\theta+\Delta\theta}$ denote the same model with LoRA adapter parameters $\Delta\theta$~\citep{hu2022lora}. We designate a \emph{verify token} $v$ (the "?" token), and two \emph{judgment tokens}: "$+$" (positive) and "$-$" (negative). We use text tokens to keep the output aligned with the base model's pre-trained vocabulary distribution \citep{xiong_implementation_2024}. 

\textbf{Scoring procedure.}
As shown in Figure \ref{fig:overview}, after agent $a_j$ completes generation, the accumulated KV cache $(\mathbf{K}, \mathbf{V})$ already exists as a byproduct of the generation process. \method scores the trajectory by processing a single verify token with the LoRA adapter activated:
\begin{equation}
    \mathbf{z} = f_{\theta+\Delta\theta}(v \mid \mathbf{K}, \mathbf{V}), \qquad
    s_j = \pplus = \text{softmax}\big([\mathbf{z}_{[-]},\; \mathbf{z}_{[+]}]\big)_1
    \label{eq:score}
\end{equation}
The LoRA adapter learns to interpret the base model's KV cache, which encodes position information, attention patterns, and hidden state interactions of the full trajectory, and outputs a quality judgment through the verify token's logits. Crucially, the KV cache is produced by the base model with the adapter \emph{disabled} during normal generation, so \method introduces no overhead to the generation process itself.

\textbf{Cost analysis.}
The verify-token forward pass processes a single query token against $L$ cached positions across $N_L$ layers:
\begin{equation}
    F_{\text{score}}^{\text{kv}} = N_L \cdot (c_{\text{attn}} \cdot d \cdot L + c_{\text{ffn}} \cdot d^2) = O(d \cdot L)
    \label{eq:cost_kv}
\end{equation}
A text-based PRM must re-encode the full trajectory from scratch, costing $F_{\text{score}}^{\text{text}} = O(d \cdot L^2)$ (Equation~\ref{eq:flops_forward}). The per-call speedup is:
\begin{equation}
    \frac{F_{\text{score}}^{\text{text}}}{F_{\text{score}}^{\text{kv}}} = \frac{O(d \cdot L^2)}{O(d \cdot L)} = O(L)
    \label{eq:speedup}
\end{equation}
For typical MAS trajectories ($L \approx 5{,}000$ tokens), this yields ${\sim}5{,}000\times$ fewer FLOPs per scoring call.
The same factor-of-$L$ reduction extends to peak activation memory: Text-PRM materializes $O(L^2)$ attention maps per layer, while \method requires only $O(L)$.
In test-time search, where the PRM is invoked $O(W \cdot D)$ times, this per-call reduction makes the total scoring cost negligible relative to generation, eliminating the computational bottleneck identified in Section~\ref{sec:prelim}.
This instantiation corresponds to $k\!=\!1$ in the readout framework, achieving the maximum theoretical speedup.

\textbf{Inference.}
\method is a drop-in replacement for text-based PRMs in all search algorithms described in Section~\ref{sec:prelim}: beam search, MCTS, and weighted majority voting. The scoring mechanism changes: \method reads the pre-existing KV cache instead of re-encoding text, but the search algorithms themselves remain unchanged.

\subsection{Training}
\label{sec:training}

\textbf{Training data generation.}
We generate training labels using Monte Carlo Tree Search (MCTS)~\citep{feng2024tsllm} over the multi-agent reasoning process. For each training question, we run $R$ MCTS rollouts with $C$ candidates per agent step. Terminal trajectories are evaluated by exact-match against the ground truth, producing binary rewards. Backpropagation through the search tree assigns Q-values $y \in [-1, 1]$ to each intermediate agent step.
Training data is generated separately for each (topology, execution mode) combination. Each (trajectory, agent step) pair produces one training sample $(\mathbf{x}_j, y_j)$.

\textbf{\method training.}
For each sample $(\mathbf{x}, y)$, the base model (adapter disabled) processes $\mathbf{x}$ to produce KV cache $(\mathbf{K}, \mathbf{V})$ without gradients.
The model (adapter active) processes the verify token conditioned on $(\mathbf{K}, \mathbf{V})$, producing score $\pplus$ via Equation~\ref{eq:score}.
The loss is:
\begin{equation}
    \mathcal{L}_{\text{KV}} = \text{MSE}\!\left(\pplus,\; \frac{y + 1}{2}\right)
    \label{eq:loss}
\end{equation}
Only the LoRA parameters $\Delta\theta$ are updated. Gradients flow only through the single verify-token forward pass. Algorithm~\ref{alg:training} summarizes the procedure.
\rev{The KV cache is generated on the fly rather than stored offline; the resulting training cost is analyzed in Appendix~\ref{app:training_cost}.}

\begin{algorithm}[t]
\caption{\method Training}
\label{alg:training}
\small 
\begin{algorithmic}[1]
\REQUIRE Training set $\mathcal{D} = \{(\mathbf{x}_i, y_i)\}$ with trajectories and MCTS Q-values, base model $f_\theta$, LoRA parameters $\Delta\theta$, learning rate $\eta$
\FOR{each mini-batch $\mathcal{B} \subset \mathcal{D}$}
    \FOR{each $(\mathbf{x}, y) \in \mathcal{B}$}
        \STATE $(\mathbf{K}, \mathbf{V}) \leftarrow f_\theta(\mathbf{x})$ \hfill \textit{// Encode trajectory with adapter off (no grad)}
        \STATE $\mathbf{z} \leftarrow f_{\theta+\Delta\theta}(v \mid \mathbf{K}, \mathbf{V})$ \hfill \textit{// Forward verify token with adapter on}
        \STATE $\pplus \leftarrow \text{softmax}\big([\mathbf{z}_{[-]},\; \mathbf{z}_{[+]}]\big)_1$
        \STATE $\mathcal{L} \leftarrow \text{MSE}\!\big(\pplus,\; (y + 1)/2\big)$
    \ENDFOR
    \STATE $\Delta\theta \leftarrow \Delta\theta - \eta\, \nabla_{\Delta\theta}\, \frac{1}{|\mathcal{B}|}\sum \mathcal{L}$ \hfill \textit{// Update only LoRA parameters}
\ENDFOR
\end{algorithmic}
\end{algorithm}

\section{Experiments}
\label{sec:experiments}

\subsection{Setup}
\label{sec:setup}

\textbf{Datasets and metrics.}
We evaluate on four mathematical reasoning benchmarks: MATH~\citep{hendrycks2021math}, GSM8K~\citep{cobbe2021gsm8k}, AIME 2024, and AIME 2025, reporting exact-match accuracy with numeric equivalence checking.
We evaluate three search algorithms: step-level beam search (SBS), MCTS, and weighted majority voting, with per-dataset configurations reported in Tables~\ref{tab:math_aime24}--\ref{tab:gsm8k_aime25}. Scoring cost is normalized to Text-PRM $= 1\times$.
Both \method and Text-PRM are trained with LoRA ($r\!=\!256$, $\alpha\!=\!32$) on MCTS-generated labels ($R\!=\!64$ rollouts, $C\!=\!4$ candidates per step) from the MATH training split. Full hyperparameters and implementation details are in Appendix~\ref{app:implementation}.

\textbf{Models and MAS configurations.}
We use three scales of the Qwen3 family~\citep{yang2025qwen3}: 0.6B, 4B, and 8B, with thinking mode disabled, under two representative multi-agent topologies: \textbf{Sequential} (Reader $\to$ Planner $\to$ Solver $\to$ Verifier) and \textbf{Hierarchical} (Math/Science/Code Agents $\to$ Task Summarizer).
\rev{To test generalization across model families, we additionally evaluate Llama-3.2-3B-Instruct~\citep{grattafiori2024llama3herdmodels} and Ministral-3-3B-Instruct~\citep{liu2026ministral3}.}

\textbf{Baselines.}
We compare \method against three scoring strategies: (1)~\textbf{No PRM} (Random Sampling / Majority Voting), where the MAS generates without process reward guidance; (2)~\textbf{Policy Log-prob}, which ranks trajectories by the generator's own token-level log-probabilities at negligible additional cost; and (3)~\textbf{Text-PRM}, a LoRA adapter trained with the same architecture, data, and objective as \method (Equation~\ref{eq:loss}), but scoring by re-encoding the full trajectory $[\mathbf{x}; v]$ in a single forward pass with the adapter active on all tokens at $O(dL^2)$ cost per call. Text-PRM training requires gradient checkpointing due to full-sequence gradient computation.

\subsection{Main Results}
\label{sec:results_main}


\textbf{KV-PRM outperforms Text-PRM across a broad range of settings while reducing scoring cost by 3 orders of magnitude.}
Table~\ref{tab:math_aime24} and Table~\ref{tab:gsm8k_aime25} show that KV-PRM achieves this accuracy-efficiency trade-off with dramatically lower scoring FLOPs per call, with relative cost reductions ranging from roughly $9\times 10^2$ to $4.9\times 10^3\times$ compared to Text-PRM, depending on the benchmark, model scale, and search configuration.
The accuracy results show that replacing full-text re-encoding with single-token KV-cache readout improves, or at least preserves, scorer quality while making verification far cheaper.


\textbf{KV-PRM significantly reduces the inference latency by 1-2 orders of magnitude.}
We validate the theoretical complexity reduction with wall-clock latency measurements on an NVIDIA GH200 GPU. As shown in the left panel of Figure~\ref{fig:latency}. KV-PRM consistently delivers substantial per-sequence speedups over Text-PRM, ranging from 15$\times$ to 37$\times$ depending on model scale and sequence length. The gap widens as sequences become longer and models become larger, in line with the expected $O(L)$ vs.\ $O(L^2)$ scoring-cost difference. At 8B and $L=4096$, a single KV-PRM scoring call takes only 4.6\,ms, compared with 172.0\,ms for Text-PRM. This confirms that KV-cache transfer translates the theoretical FLOP savings into large practical reductions in end-to-end scorer latency.

\textbf{KV-PRM significantly reduces the memory consumption during inference by 1-2 orders of magnitude.}
As shown in the right panel of Figure~\ref{fig:latency}, \method reduces the average memory increase per sequence by up to $34.2\times$, relative to Text-PRM under matched batch sizes. This gain comes from avoiding full-sequence re-encoding and the associated attention-map materialization, allowing KV-PRM to score trajectories with substantially lower activation overhead. In practice, the smaller memory footprint enables larger verification batches on the same hardware, providing a complementary systems advantage beyond latency alone.


\begin{table}[t]
\caption{Test-time search accuracy (\%) and relative scoring cost on \textbf{MATH} and \textbf{AIME 2024} under sequential multi-agent execution. Scoring cost normalized to Text-PRM $= 1\times$; Policy Log-prob cost is negligible (---); \method cost $= 1/L$ where $L$ is the maximum mean per-step trajectory length.}
\label{tab:math_aime24}
\centering
\resizebox{\textwidth}{!}{
\begin{tabular}{ll l cc cc cc cc cc c}
\toprule
& & & \multicolumn{6}{c}{\textbf{MATH}} & \multicolumn{4}{c}{\textbf{AIME 2024}} & \\
\cmidrule(lr){4-9} \cmidrule(lr){10-13}
& & & \multicolumn{2}{c}{\textbf{Qwen3-0.6B}} & \multicolumn{2}{c}{\textbf{Qwen3-4B}} & \multicolumn{2}{c}{\textbf{Qwen3-8B}} & \multicolumn{2}{c}{\textbf{Qwen3-4B}} & \multicolumn{2}{c}{\textbf{Qwen3-8B}} & \\
\cmidrule(lr){4-5} \cmidrule(lr){6-7} \cmidrule(lr){8-9} \cmidrule(lr){10-11} \cmidrule(lr){12-13}
\textbf{Search} & \textbf{Config} & \textbf{Scorer} & Acc & Cost & Acc & Cost & Acc & Cost & Acc & Cost & Acc & Cost & \textbf{Avg.} \\
\midrule
Random Sampling & $n\!=\!1$ & --- & 28.00 & & 54.15 & & 56.70 & & 20.00 & & 20.00 & & 35.77 \\
Majority Voting & $n\!=\!40$ & --- & 34.65 & & 59.20 & & 60.55 & & 36.67 & & 33.33 & & 44.88 \\
\midrule
Beam Search & $W\!=\!1$ & Policy Log-prob & 32.50 & --- & 58.70 & --- & 63.70 & --- & 23.33 & --- & 23.33 & --- & 40.31 \\
            &             & Text-PRM & 32.75 & $1\times$ & 62.00 & $1\times$ & 66.62 & $1\times$ & 26.67 & $1\times$ & 20.00 & $1\times$ & 41.61 \\
            &             & \cellcolor{gray!15}\method & \cellcolor{gray!15}31.90 & \cellcolor{gray!15}\textbf{$\frac{1}{2409}$} & \cellcolor{gray!15}\textbf{65.55} & \cellcolor{gray!15}\textbf{$\frac{1}{2543}$} & \cellcolor{gray!15}\textbf{67.02} & \cellcolor{gray!15}\textbf{$\frac{1}{2782}$} & \cellcolor{gray!15}20.00 & \cellcolor{gray!15}\textbf{$\frac{1}{2007}$} & \cellcolor{gray!15}\textbf{30.00} & \cellcolor{gray!15}\textbf{$\frac{1}{2367}$} & \cellcolor{gray!15}\textbf{42.89} \\
\midrule
MCTS & $n\!=\!50$ & Policy Log-prob & 29.40 & --- & 63.90 & --- & 61.90 & --- & 20.00 & --- & 23.33 & --- & 39.71 \\
            &             & Text-PRM & 35.55 & $1\times$ & 65.95 & $1\times$ & 68.92 & $1\times$ & 30.00 & $1\times$ & 36.67 & $1\times$ & 47.42 \\
            &             & \cellcolor{gray!15}\method & \cellcolor{gray!15}35.10 & \cellcolor{gray!15}\textbf{$\frac{1}{3598}$} & \cellcolor{gray!15}\textbf{67.15} & \cellcolor{gray!15}\textbf{$\frac{1}{3771}$} & \cellcolor{gray!15}\textbf{69.57} & \cellcolor{gray!15}\textbf{$\frac{1}{4367}$} & \cellcolor{gray!15}\textbf{33.33} & \cellcolor{gray!15}\textbf{$\frac{1}{3652}$} & \cellcolor{gray!15}\textbf{40.00} & \cellcolor{gray!15}\textbf{$\frac{1}{3944}$} & \cellcolor{gray!15}\textbf{49.03} \\
\midrule
\multirow{6}{*}{Weighted Voting} & $n\!=\!10$ & Policy Log-prob & 33.75 & --- & 58.45 & --- & 59.95 & --- & 30.00 & --- & 33.33 & --- & 43.10 \\
            &             & Text-PRM & 34.20 & $1\times$ & 58.75 & $1\times$ & 60.51 & $1\times$ & 36.67 & $1\times$ & 23.33 & $1\times$ & 42.69 \\
            &             & \cellcolor{gray!15}\method & \cellcolor{gray!15}\textbf{34.40} & \cellcolor{gray!15}\textbf{$\frac{1}{2444}$} & \cellcolor{gray!15}\textbf{59.35} & \cellcolor{gray!15}\textbf{$\frac{1}{2749}$} & \cellcolor{gray!15}60.46 & \cellcolor{gray!15}\textbf{$\frac{1}{3137}$} & \cellcolor{gray!15}33.33 & \cellcolor{gray!15}\textbf{$\frac{1}{2360}$} & \cellcolor{gray!15}\textbf{33.33} & \cellcolor{gray!15}\textbf{$\frac{1}{2575}$} & \cellcolor{gray!15}\textbf{44.17} \\
\cmidrule{2-14}
 & $n\!=\!200$ & Policy Log-prob & 35.35 & --- & 59.40 & --- & 60.85 & --- & 36.67 & --- & 33.33 & --- & 45.12 \\
            &             & Text-PRM & 36.05 & $1\times$ & 59.95 & $1\times$ & 61.26 & $1\times$ & 36.67 & $1\times$ & 33.33 & $1\times$ & 45.45 \\
            &             & \cellcolor{gray!15}\method & \cellcolor{gray!15}\textbf{36.15} & \cellcolor{gray!15}\textbf{$\frac{1}{2441}$} & \cellcolor{gray!15}\textbf{60.15} & \cellcolor{gray!15}\textbf{$\frac{1}{2721}$} & \cellcolor{gray!15}61.06 & \cellcolor{gray!15}\textbf{$\frac{1}{3086}$} & \cellcolor{gray!15}\textbf{36.67} & \cellcolor{gray!15}\textbf{$\frac{1}{2298}$} & \cellcolor{gray!15}\textbf{36.67} & \cellcolor{gray!15}\textbf{$\frac{1}{2539}$} & \cellcolor{gray!15}\textbf{46.14} \\
\bottomrule
\end{tabular}}
\end{table}

\begin{table}[t]
\caption{Test-time search accuracy (\%) and relative scoring cost on \textbf{GSM8K} and \textbf{AIME 2025} under sequential multi-agent execution. Same format as Table~\ref{tab:math_aime24}.}
\label{tab:gsm8k_aime25}
\centering
\resizebox{\textwidth}{!}{
\begin{tabular}{ll l cc cc cc cc cc c}
\toprule
& & & \multicolumn{6}{c}{\textbf{GSM8K}} & \multicolumn{4}{c}{\textbf{AIME 2025}} & \\
\cmidrule(lr){4-9} \cmidrule(lr){10-13}
& & & \multicolumn{2}{c}{\textbf{Qwen3-0.6B}} & \multicolumn{2}{c}{\textbf{Qwen3-4B}} & \multicolumn{2}{c}{\textbf{Qwen3-8B}} & \multicolumn{2}{c}{\textbf{Qwen3-4B}} & \multicolumn{2}{c}{\textbf{Qwen3-8B}} & \\
\cmidrule(lr){4-5} \cmidrule(lr){6-7} \cmidrule(lr){8-9} \cmidrule(lr){10-11} \cmidrule(lr){12-13}
\textbf{Search} & \textbf{Config} & \textbf{Scorer} & Acc & Cost & Acc & Cost & Acc & Cost & Acc & Cost & Acc & Cost & \textbf{Avg.} \\
\midrule
Random Sampling & $n\!=\!1$ & --- & 50.80 & & 91.28 & & 91.43 & & 16.67 & & 20.00 & & 54.04 \\
Majority Voting & $n\!=\!40$ & --- & 62.77 & & 93.10 & & 93.03 & & 20.00 & & 16.67 & & 57.11 \\
\midrule
Beam Search & $W\!=\!10$ & Policy Log-prob & 52.99 & --- & 90.98 & --- & 93.40 & --- & 13.33 & --- & 16.67 & --- & 53.47 \\
            &             & Text-PRM & 58.68 & $1\times$ & 93.78 & $1\times$ & 95.00 & $1\times$ & 26.67 & $1\times$ & 26.67 & $1\times$ & 60.16 \\
            &             & \cellcolor{gray!15}\method & \cellcolor{gray!15}58.00 & \cellcolor{gray!15}\textbf{$\frac{1}{1341}$} & \cellcolor{gray!15}92.80 & \cellcolor{gray!15}\textbf{$\frac{1}{1311}$} & \cellcolor{gray!15}\textbf{95.07} & \cellcolor{gray!15}\textbf{$\frac{1}{1295}$} & \cellcolor{gray!15}\textbf{26.67} & \cellcolor{gray!15}\textbf{$\frac{1}{4266}$} & \cellcolor{gray!15}\textbf{30.00} & \cellcolor{gray!15}\textbf{$\frac{1}{4392}$} & \cellcolor{gray!15}\textbf{60.51} \\
\midrule
MCTS & $n\!=\!30$ & Policy Log-prob & 52.92 & --- & 89.99 & --- & 91.13 & --- & 13.33 & --- & 16.67 & --- & 52.81 \\
            &             & Text-PRM & 57.77 & $1\times$ & 93.33 & $1\times$ & 95.15 & $1\times$ & 23.33 & $1\times$ & 23.33 & $1\times$ & 58.58 \\
            &             & \cellcolor{gray!15}\method & \cellcolor{gray!15}54.74 & \cellcolor{gray!15}\textbf{$\frac{1}{1357}$} & \cellcolor{gray!15}92.87 & \cellcolor{gray!15}\textbf{$\frac{1}{1386}$} & \cellcolor{gray!15}94.62 & \cellcolor{gray!15}\textbf{$\frac{1}{1402}$} & \cellcolor{gray!15}\textbf{23.33} & \cellcolor{gray!15}\textbf{$\frac{1}{4877}$} & \cellcolor{gray!15}\textbf{26.67} & \cellcolor{gray!15}\textbf{$\frac{1}{4840}$} & \cellcolor{gray!15}58.45 \\
\midrule
\multirow{6}{*}{Weighted Voting} & $n\!=\!20$ & Policy Log-prob & 61.64 & --- & 92.72 & --- & 92.87 & --- & 16.67 & --- & 16.67 & --- & 56.11 \\
            &             & Text-PRM & 62.40 & $1\times$ & 93.18 & $1\times$ & 93.18 & $1\times$ & 13.33 & $1\times$ & 16.67 & $1\times$ & 55.75 \\
            &             & \cellcolor{gray!15}\method & \cellcolor{gray!15}\textbf{63.61} & \cellcolor{gray!15}\textbf{$\frac{1}{908}$} & \cellcolor{gray!15}92.87 & \cellcolor{gray!15}\textbf{$\frac{1}{1006}$} & \cellcolor{gray!15}93.03 & \cellcolor{gray!15}\textbf{$\frac{1}{1036}$} & \cellcolor{gray!15}\textbf{20.00} & \cellcolor{gray!15}\textbf{$\frac{1}{3402}$} & \cellcolor{gray!15}\textbf{20.00} & \cellcolor{gray!15}\textbf{$\frac{1}{3415}$} & \cellcolor{gray!15}\textbf{57.90} \\
\cmidrule{2-14}
 & $n\!=\!100$ & Policy Log-prob & 63.00 & --- & 92.80 & --- & 92.95 & --- & 16.67 & --- & 16.67 & --- & 56.42 \\
            &             & Text-PRM & 63.38 & $1\times$ & 93.18 & $1\times$ & 93.25 & $1\times$ & 20.00 & $1\times$ & 16.67 & $1\times$ & 57.30 \\
            &             & \cellcolor{gray!15}\method & \cellcolor{gray!15}\textbf{63.76} & \cellcolor{gray!15}\textbf{$\frac{1}{934}$} & \cellcolor{gray!15}\textbf{93.18} & \cellcolor{gray!15}\textbf{$\frac{1}{997}$} & \cellcolor{gray!15}\textbf{93.25} & \cellcolor{gray!15}\textbf{$\frac{1}{1025}$} & \cellcolor{gray!15}\textbf{23.33} & \cellcolor{gray!15}\textbf{$\frac{1}{3428}$} & \cellcolor{gray!15}\textbf{16.67} & \cellcolor{gray!15}\textbf{$\frac{1}{3380}$} & \cellcolor{gray!15}\textbf{58.04} \\
\bottomrule
\end{tabular}}
\end{table}

\subsection{Analysis}
\label{sec:analysis}

\begin{figure}[t]
\centering
\begin{subfigure}[t]{0.45\linewidth}
    \centering
    \includegraphics[width=\linewidth]{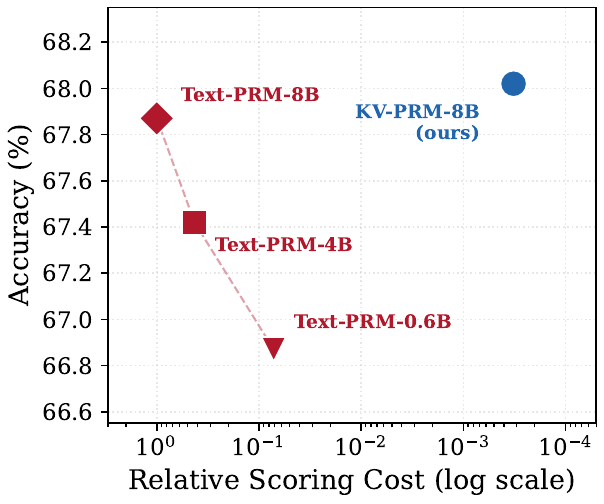}
    \label{fig:cost_vs_acc}
\end{subfigure}
\hfill
\begin{subfigure}[t]{0.45\linewidth}
    \centering
    \includegraphics[width=\linewidth]{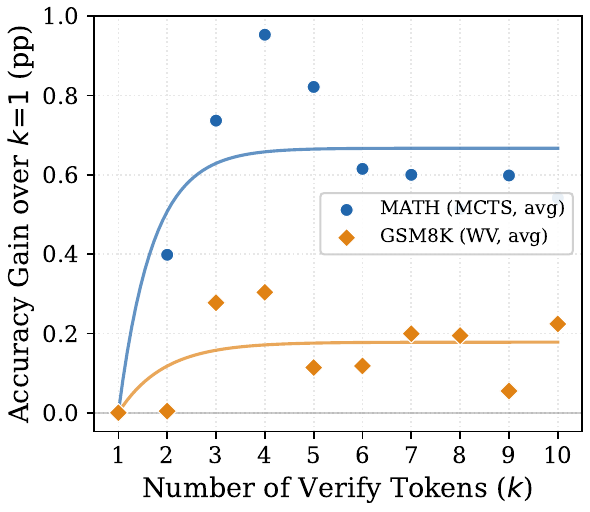}
    \label{fig:scaling}
\end{subfigure}
\vspace{-5mm}
\caption{Analysis of \method efficiency and readout depth. \emph{Left}: Scaling up the Text-PRM from 0.6B to 8B yields diminishing returns; \method surpasses all sizes at ${\sim}10^3\times$ lower cost. \emph{Right}: Accuracy gain over $k\!=\!1$ as a function of the number of verify tokens $k$ averaged across search hyperparameters. Fitted curves $a(1-e^{-c(k-1)})$ confirm exponentially diminishing returns.}
\label{fig:analysis}
\end{figure}

\textbf{Can smaller Text-PRMs close the efficiency gap?}
A natural strategy for reducing Text-PRM scoring cost is to use a smaller verifier model, which is a common practice in the literature~\citep{snell2024scaling}.
We evaluate this on the MATH dataset, using Qwen3-8B as the base generator and scoring the same search trees with Text-PRMs of sizes 0.6B, 4B, and 8B alongside \method of size 8B.
Figure~\ref{fig:analysis}(Left) plots accuracy against relative scoring cost per call for SBS $W\!=\!2$.
Scaling up the Text-PRM from 0.6B to 4B to 8B improves accuracy, but with diminishing returns and increasing cost.
\method achieves the \emph{highest} accuracy (68.0\%), surpassing all Text-PRM sizes, at ${\sim}3{,}000\times$ lower cost than Text-PRM-8B, demonstrating that KV-cache transfer is a fundamentally more efficient strategy than reducing the size of the PRM.

\textbf{What happens with more verify tokens ($k > 1$)?}
Theorem~\ref{thm:diminishing} makes two concrete predictions: (a)~the marginal information gain $\Delta I_k$ from each additional verify token decays exponentially in $k$ at rate $\alpha \cdot n_h N_L$ (Equation~\ref{eq:diminishing}), and (b)~a single token ($k\!=\!1$) already captures most extractable reward information (Equation~\ref{eq:k1_optimality}).
Figure~\ref{fig:analysis}(Right) tests these predictions on Qwen3-0.6B ($n_h \!\cdot\! N_L \!=\! 448$), plotting accuracy gain over $k\!=\!1$ averaged across search hyperparameters.
Both curves are well fit by $a(1\!-\!e^{-c(k-1)})$, consistent with the exponential decay form $e^{-\alpha \cdot n_h N_L \cdot (k-1)}$ predicted by Theorem~\ref{thm:diminishing}(a). The rapid saturation validates the $k\!=\!1$ near-optimality bound (Theorem~\ref{thm:diminishing}b): even at the smallest model scale, a single verify token captures most of the available reward signal.
By the cost, information tradeoff (Theorem~\ref{thm:diminishing}c), the information gained per additional FLOP decays as $O(e^{-\alpha \cdot n_h N_L \cdot (k-1)} / (dL))$, so even at $k\!=\!4$ the total scoring cost of $4 \cdot O(dL)$ remains ${\sim}10^3\times$ cheaper than text re-encoding at $O(dL^2)$.


\begin{figure}[t]
\centering
\includegraphics[width=0.95\linewidth]{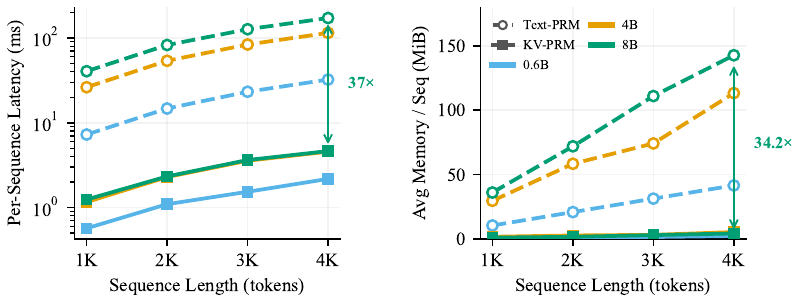}
\vspace{-2mm}
\caption{Wall-clock efficiency on NVIDIA GH200 120GB GPU. \emph{Left}: Per-sequence scoring latency (log scale). \method (solid) is $1$--$2$ orders of magnitude faster than Text-PRM (dashed), reaching $37\times$ at 8B ($L\!=\!4{,}096$). \emph{Right}: Per-sequence average memory increase during PRM inference. \method reduces memory by up to $34.2\times$, enabling substantially larger batches on the same hardware.}
\label{fig:latency}
\end{figure}


\textbf{Does KV-PRM work on more MAS topologies?}
We additionally evaluate \method under a hierarchical multi-agent topology.
Table~\ref{tab:hier} reports results across all four datasets.
\method matches or outperforms Text-PRM in the majority of search configurations. 
On MATH, the advantage is consistent but modest ($+0.5$--$0.8$ pp), while on AIME benchmarks the gaps are more pronounced (up to $+6.7$ pp), reflecting greater headroom for scorer quality on harder problems.
These results confirm that the KV-cache scoring mechanism works effectively for non-sequential MAS topologies.

\rev{\textbf{Does KV-PRM generalize to other model families?}
Table~\ref{tab:families} reports results on Llama-3.2-3B and Ministral-3-3B.
The native \method surpasses Text-PRM in average accuracy in five of the six backbone--search combinations and trails it by only $0.11$ pp in the remaining one, while its scoring cost remains three orders of magnitude lower, as on Qwen3 (Appendix~\ref{app:families}).
This confirms that KV-cache transfer is not specific to the Qwen3 architecture.}

\rev{\textbf{Can KV-PRM score a generator of a different architecture?}
To relax the requirement of a shared architecture, we train a lightweight adapter that converts the generator's KV cache into the cache geometry of a frozen \method of another family, which then scores it with a single verify token without re-prefilling the context (details in Appendix~\ref{app:hetero}).
As shown in Table~\ref{tab:families}, such cross-architecture variants trail the native \method by $2.1$--$4.6$ pp in average accuracy under beam search and MCTS, yet still surpass Policy Log-prob in these settings for both generators, demonstrating the feasibility of extending \method to heterogeneous generator--verifier pairs.
Under weighted voting, where all scorers lie within $1.4$ pp of each other, they fall slightly below Policy Log-prob.}

\rev{\textbf{Does KV-PRM transfer beyond math?}
We apply the math-trained \method and Text-PRM, without retraining, to guide test-time search on HumanEval~\citep{chen2021evaluating} and HumanEval+~\citep{evalplus}.
As shown in Table~\ref{tab:coding}, \method raises HumanEval accuracy from $71.34\%$ (random sampling) to $80.49\%$ with MCTS, outperforms Policy Log-prob in all settings, and stays within $0.61$ pp of, or above, Text-PRM.}

\rev{\textbf{Additional analyses.}
Comparisons with a last-hidden-state value head and with end-only verification, as well as score calibration and training cost, are provided in Appendix~\ref{app:additional}.}

\begin{table*}[t]
\centering

\begin{minipage}[t]{0.47\textwidth}
\centering
\caption{Hierarchical MAS accuracy (\%) across datasets and model scales. \method matches or exceeds Text-PRM on all dataset--model combinations. $\Delta$ = \method $-$ Text-PRM.}
\label{tab:hier}
\small
\adjustbox{max width=\linewidth}{
\begin{tabular}{ll ccc}
\toprule
\textbf{Dataset} & \textbf{Model} & \textbf{Text-PRM} & \cellcolor{gray!15}\textbf{\method} & \textbf{$\Delta$} \\
\midrule
\multirow{2}{*}{MATH}
& Qwen3-4B & 58.7 & \cellcolor{gray!15}\textbf{59.2} & $+0.5$ \\
& Qwen3-8B & 60.2 & \cellcolor{gray!15}\textbf{61.0} & $+0.8$ \\
\midrule
\multirow{2}{*}{GSM8K}
& Qwen3-4B & 93.7 & \cellcolor{gray!15}\textbf{93.9} & $+0.2$ \\
& Qwen3-8B & 94.2 & \cellcolor{gray!15}\textbf{94.5} & $+0.3$ \\
\midrule
\multirow{2}{*}{AIME 2024}
& Qwen3-4B & 23.3 & \cellcolor{gray!15}\textbf{26.7} & $+3.3$ \\
& Qwen3-8B & 23.3 & \cellcolor{gray!15}\textbf{30.0} & $+6.7$ \\
\midrule
\multirow{2}{*}{AIME 2025}
& Qwen3-4B & 23.3 & \cellcolor{gray!15}\textbf{26.7} & $+3.3$ \\
& Qwen3-8B & 16.7 & \cellcolor{gray!15}\textbf{20.0} & $+3.3$ \\
\bottomrule
\end{tabular}
}
\end{minipage}
\hfill
\begin{minipage}[t]{0.51\textwidth}
\centering
\caption{\steer proof-of-concept: accuracy with gradient-based KV-cache optimization at agent handoffs, without any search. This is structurally impossible with text-based PRMs.}
\label{tab:steering_main}
\small
\adjustbox{max width=\linewidth}{
\begin{tabular}{ll ccc}
\toprule
\textbf{Dataset} & \textbf{Model} & \textbf{No steering} & \cellcolor{gray!15}\textbf{\steer} & \textbf{$\Delta$} \\
\midrule
\multirow{2}{*}{MATH}
& Qwen3-4B & 60.23 & \cellcolor{gray!15}\textbf{61.84} & $+1.61$ \\
& Qwen3-8B & 60.79 & \cellcolor{gray!15}\textbf{62.00} & $+1.21$ \\
\midrule
\multirow{2}{*}{GSM8K}
& Qwen3-4B & 91.81 & \cellcolor{gray!15}\textbf{92.42} & $+0.61$ \\
& Qwen3-8B & 93.86 & \cellcolor{gray!15}\textbf{94.09} & $+0.23$ \\
\midrule
\multirow{2}{*}{AIME 2024}
& Qwen3-4B & 10.00 & \cellcolor{gray!15}\textbf{13.33} & $+3.33$ \\
& Qwen3-8B & 13.33 & \cellcolor{gray!15}\textbf{16.67} & $+3.33$ \\
\midrule
\multirow{2}{*}{AIME 2025}
& Qwen3-4B & 10.00 & \cellcolor{gray!15}{10.00} & $+0.00$ \\
& Qwen3-8B & 13.33 & \cellcolor{gray!15}\textbf{16.67} & $+3.33$ \\
\bottomrule
\end{tabular}
}
\end{minipage}

\end{table*}

\begin{table}[t]
\rev{%
\caption{\rev{Test-time search accuracy (\%) on \textbf{Llama-3.2-3B-Instruct} and \textbf{Ministral-3-3B-Instruct}. \method: native verifier; \method-$X$: verifier of family $X$ reading the generator's KV cache through a cross-architecture adapter.}}
\label{tab:families}
\centering
\resizebox{\textwidth}{!}{
\begin{tabular}{l l ccccc l ccccc}
\toprule
& \multicolumn{6}{c}{\textbf{Generator: Llama-3.2-3B}} & \multicolumn{6}{c}{\textbf{Generator: Ministral-3-3B}} \\
\cmidrule(lr){2-7} \cmidrule(lr){8-13}
\textbf{Search} & \textbf{Scorer} & MATH & GSM8K & AIME 24 & AIME 25 & \textbf{Avg.} & \textbf{Scorer} & MATH & GSM8K & AIME 24 & AIME 25 & \textbf{Avg.} \\
\midrule
Random Sampling ($n\!=\!1$) & --- & 45.35 & 74.07 & 13.33 & 0.00 & 33.19 & --- & 50.25 & 87.49 & 10.00 & 6.67 & 38.60 \\
Majority Voting ($n\!=\!160$) & --- & 54.85 & 84.53 & 16.67 & 0.00 & 39.01 & --- & 57.45 & 92.87 & 23.33 & 20.00 & 48.41 \\
\midrule
\multirow{5}{*}{\shortstack[l]{Beam Search\\$W\!=\!4$}}
& Policy Log-prob & 52.80 & 80.82 & 10.00 & 0.00 & 35.90 & Policy Log-prob & 56.10 & 90.83 & 16.67 & 16.67 & 45.06 \\
& Text-PRM & 58.00 & 86.73 & 13.33 & 0.00 & 39.52 & Text-PRM & 63.45 & 92.80 & 26.67 & 20.00 & 50.73 \\
& \method-Qwen3 & 52.55 & 85.37 & 6.67 & 0.00 & 36.15 & \method-Qwen3 & 60.60 & 91.96 & 23.33 & 16.67 & 48.14 \\
& \method-Ministral & 54.65 & 85.29 & 6.67 & 0.00 & 36.65 & \method-Llama & 58.45 & 91.66 & 23.33 & 20.00 & 48.36 \\
& \cellcolor{gray!15}\method & \cellcolor{gray!15}57.95 & \cellcolor{gray!15}86.35 & \cellcolor{gray!15}\textbf{13.33} & \cellcolor{gray!15}\textbf{0.00} & \cellcolor{gray!15}39.41 & \cellcolor{gray!15}\method & \cellcolor{gray!15}\textbf{63.50} & \cellcolor{gray!15}\textbf{93.03} & \cellcolor{gray!15}20.00 & \cellcolor{gray!15}\textbf{26.67} & \cellcolor{gray!15}\textbf{50.80} \\
\midrule
\multirow{5}{*}{\shortstack[l]{MCTS\\$n\!=\!160$}}
& Policy Log-prob & 52.25 & 78.39 & 6.67 & 0.00 & 34.33 & Policy Log-prob & 55.85 & 91.51 & 20.00 & 16.67 & 46.01 \\
& Text-PRM & 59.35 & 88.10 & 16.67 & 0.00 & 41.03 & Text-PRM & 65.95 & 93.48 & 20.00 & 23.33 & 50.69 \\
& \method-Qwen3 & 55.70 & 86.88 & 6.67 & 0.00 & 37.31 & \method-Qwen3 & 61.80 & 93.10 & 20.00 & 20.00 & 48.73 \\
& \method-Ministral & 57.00 & 87.26 & 10.00 & 0.00 & 38.57 & \method-Llama & 60.70 & 93.03 & 20.00 & 23.33 & 49.26 \\
& \cellcolor{gray!15}\method & \cellcolor{gray!15}59.25 & \cellcolor{gray!15}\textbf{88.25} & \cellcolor{gray!15}\textbf{16.67} & \cellcolor{gray!15}\textbf{3.33} & \cellcolor{gray!15}\textbf{41.87} & \cellcolor{gray!15}\method & \cellcolor{gray!15}65.90 & \cellcolor{gray!15}92.87 & \cellcolor{gray!15}\textbf{23.33} & \cellcolor{gray!15}\textbf{23.33} & \cellcolor{gray!15}\textbf{51.36} \\
\midrule
\multirow{5}{*}{\shortstack[l]{Weighted Voting\\$n\!=\!200$}}
& Policy Log-prob & 55.05 & 83.55 & 16.67 & 0.00 & 38.82 & Policy Log-prob & 57.45 & 93.03 & 23.33 & 20.00 & 48.45 \\
& Text-PRM & 54.65 & 85.67 & 13.33 & 0.00 & 38.41 & Text-PRM & 57.90 & 93.40 & 23.33 & 20.00 & 48.66 \\
& \method-Qwen3 & 55.15 & 84.84 & 10.00 & 0.00 & 37.50 & \method-Qwen3 & 57.25 & 93.10 & 23.33 & 20.00 & 48.42 \\
& \method-Ministral & 55.35 & 85.37 & 10.00 & 0.00 & 37.68 & \method-Llama & 57.15 & 93.03 & 23.33 & 20.00 & 48.38 \\
& \cellcolor{gray!15}\method & \cellcolor{gray!15}\textbf{55.30} & \cellcolor{gray!15}\textbf{85.75} & \cellcolor{gray!15}\textbf{13.33} & \cellcolor{gray!15}\textbf{0.00} & \cellcolor{gray!15}\textbf{38.60} & \cellcolor{gray!15}\method & \cellcolor{gray!15}\textbf{57.90} & \cellcolor{gray!15}\textbf{93.48} & \cellcolor{gray!15}\textbf{23.33} & \cellcolor{gray!15}\textbf{20.00} & \cellcolor{gray!15}\textbf{48.68} \\
\bottomrule
\end{tabular}}
}
\end{table}

\begin{table}[t]
\rev{%
\caption{\rev{Test-time search accuracy (\%) on code generation with PRMs trained on MATH only.}}
\label{tab:coding}
\centering
\small
\begin{tabular}{l c ccc ccc}
\toprule
& & \multicolumn{3}{c}{\textbf{Beam Search} ($W\!=\!2$)} & \multicolumn{3}{c}{\textbf{MCTS} ($n\!=\!40$)} \\
\cmidrule(lr){3-5} \cmidrule(lr){6-8}
\textbf{Dataset} & \shortstack{\textbf{Random}\\\textbf{Sampling}} & \shortstack{Policy\\Log-prob} & Text-PRM & \cellcolor{gray!15}\method & \shortstack{Policy\\Log-prob} & Text-PRM & \cellcolor{gray!15}\method \\
\midrule
HumanEval  & 71.34 & 75.00 & 78.66 & \cellcolor{gray!15}78.05 & 76.83 & 79.27 & \cellcolor{gray!15}\textbf{80.49} \\
HumanEval+ & 62.20 & 66.46 & 70.12 & \cellcolor{gray!15}69.51 & 70.73 & 72.56 & \cellcolor{gray!15}71.95 \\
\bottomrule
\end{tabular}
}
\end{table}

\section{Discussion on Broader Applications of \method}
\label{sec:steering}
As \method scores are differentiable with respect to the KV cache, we explore whether gradient-based optimization of latent inter-agent messages~\cite{jin2026agent} can improve generation quality, a capability structurally impossible with text-based PRMs. 
\steer performs gradient ascent on the accumulated KV cache at each agent handoff to maximize the \method score before the next agent begins generation (algorithm details in Appendix~\ref{app:steering}).
Table~\ref{tab:steering_main} presents proof-of-concept results.
\steer improves accuracy on MATH by $+1.6$ pp (Qwen3-4B) and $+1.2$ pp (Qwen3-8B) without any search.
On AIME benchmarks with Qwen3-8B, steering yields $+3.3$ pp gains on both AIME 2024 and AIME 2025.
While preliminary, these results demonstrate that the KV-native architecture opens a qualitatively new direction for inference-time optimization beyond search.

\section{Related Work}
\label{sec:related}
\textbf{Process reward models.}
PRMs~\cite{khalifa2025thinkprm,zhao2025genprm,zheng2025prmsurvey,kuang2025timprmverifyingmultimodalreasoning} provide step-level supervision for mathematical reasoning, pioneered by \citet{lightman2023lets} who demonstrated that step-level human feedback improves over outcome-based reward models.
Math-Shepherd~\citep{wang2024mathshepherd} automates PRM label generation using MCTS.
OmegaPRM~\citep{luo2024omegaprm} further scales automated PRM training.
GenRM~\citep{zhang2024genrm} frames verification as generation.
All existing PRMs operate on text representations; \method is the first to score via KV-cache transfer, reducing the computational cost by orders of magnitude.

\textbf{LLM-based multi-agent systems.}
Multi-agent systems have been explored in debate frameworks~\citep{du2023debate}, collaborative problem-solving~\citep{li2023camel,lin2026squidagentparallelizewiselycoordinate}, and structured pipelines~\citep{hong2024metagpt}.
MASPRM~\citep{yazdani2025masprm} introduced PRMs for multi-agent systems using text-based scoring.
We address the computational bottleneck that MASPRM's text-based approach creates in test-time search.

\textbf{KV-cache methods.}
Prior work has explored KV-cache compression~\citep{li2024kvsurvey}, eviction~\citep{xiao2023attentionsinks}, and memory management via paging~\citep{kwon2023vllm} for efficient inference.
These methods aim to reduce the \emph{size} or improve the \emph{allocation} of the KV cache; our work instead uses the KV cache as an \emph{input representation} for reward modeling, a fundamentally different application.

\textbf{Test-time compute scaling.}
Scaling test-time compute through search~\citep{snell2024scaling}, self-consistency~\citep{wang2023selfconsistency}, tree-of-thoughts~\citep{yao2023tot}, MCTS~\citep{feng2024tsllm}, and reinforcement learning for reasoning~\citep{deepseek2025r1} has shown consistent gains.
However, the cost of scoring candidates, particularly ones with long trajectories, has received limited attention. Our work directly addresses this bottleneck.

\section{Limitations and Conclusion}
\label{sec:conclusion}


We presented \method, a process reward model that scores trajectories through a single verify-token forward pass over the generator's pre-existing KV cache, reducing per-call verifier cost from $O(dL^2)$ to $O(dL)$. 
\rev{Our theoretical framework proves that the KV cache is a richer representation than decoded text for verification (Proposition~\ref{thm:repr_advantage}),} and that the marginal information gain from additional readout tokens decays exponentially, making the efficient $k\!=\!1$ design near-optimal (Theorem~\ref{thm:diminishing}). Empirically, \method matches or outperforms text-based PRMs across beam search, MCTS, and weighted voting on four benchmarks\rev{, three model families,} and two MAS topologies, while delivering up to ${\sim}5{,}000\times$ fewer scoring FLOPs, $37\times$ lower latency, and $34\times$ less memory per sequence. The differentiability of KV-cache scoring further enables \steer, a proof-of-concept for gradient-based inter-agent message optimization that is structurally impossible with text-based PRMs. 
Limitations include the requirement that generator and verifier share the same architecture\rev{, which cache adaptation only partially relaxes, the need for provider-side KV-cache access (Appendix~\ref{app:deployment})}, reliance on the Linear Representation Hypothesis in the theoretical bounds, and the preliminary nature of \steer. We hope this work encourages a shift from redundant text re-encoding toward representation reuse in verifier design.


\begin{ack}
This work was partially supported by the National Artificial Intelligence Research Resource (NAIRR) Pilot under awards NAIRR250400 and NAIRR240283, and Standing Up to POTS. 
\end{ack}

\bibliographystyle{unsrtnat} 
\bibliography{references}

\newpage
\appendix

\section{Theoretical Proofs}
\label{app:theory}

\subsection{Proof of \rev{Proposition}~\ref{thm:repr_advantage}}

\textbf{Part (a): Information sufficiency.}
The generation process defines the Markov chain $Y \to \mathbf{H} \to \mathbf{x}$: \rev{for the KV cache $\mathbf{H} = (\mathbf{K}, \mathbf{V})$ and any target variable $Y$,} text $\mathbf{x} = \mathrm{decode}(\mathbf{H})$ is a deterministic function of $\mathbf{H}$ (each token $x_t = \arg\max_v [W_{\mathrm{out}} h_t]_v$ or is sampled from $\mathrm{softmax}(W_{\mathrm{out}} h_t)$, where $h_t$ is determined by $\mathbf{H}$).
By the Data Processing Inequality~\citep[Theorem~2.8.1]{cover2006information}, for any Markov chain $Y \to \mathbf{H} \to \mathbf{x}$:
\begin{equation}
    I(\mathbf{H};\, Y) \;\geq\; I(\mathbf{x};\, Y).
\end{equation}
Equality holds if and only if $\mathbf{x}$ is a sufficient statistic of $\mathbf{H}$ for $Y$, \ie $Y \perp \mathbf{H} \mid \mathbf{x}$.
Since $\mathrm{decode}(\cdot)$ maps from $\mathbb{R}^{N_L \times L \times d}$ to $\{1, \ldots, |\mathcal{V}|\}^L$---a many-to-one projection at each position---distinct hidden states that carry different information about $Y$ are generically collapsed to the same token, so equality does not hold in general. \hfill $\square$

\textbf{Part (b): Capacity bound.}
Under Assumption~\ref{assum:lrh}, the set of hidden embeddings is
\begin{equation}
    \mathcal{H} = \left\{ \sum_{i=1}^{d} c_i s_i : c_1, \ldots, c_d \in \{0, \pm 1\} \right\},
\end{equation}
with $|\mathcal{H}| = 3^d$ (since the semantic basis is linearly independent).
For a trajectory of length $L$, the KV cache encodes $L$ such hidden states (considering the final-layer representation per position), giving $|\mathcal{H}^L| = 3^{dL}$ distinct configurations.

To represent these losslessly via text tokens from vocabulary $\mathcal{V}$, a text sequence of length $m'$ can encode at most $|\mathcal{V}|^{m'}$ distinct configurations. Lossless representation requires:
\begin{equation}
    |\mathcal{V}|^{m'} \;\geq\; 3^{dL}.
\end{equation}
Taking logarithms:
\begin{equation}
    m' \;\geq\; \frac{dL \cdot \log 3}{\log |\mathcal{V}|} \;=\; \Omega\!\left(\frac{d \cdot L}{\log |\mathcal{V}|}\right). \quad \square
\end{equation}

\subsection{Proof of Theorem~\ref{thm:readout}}

\textbf{Verification error decomposition.}
For any representation $R$ and function class $\mathcal{F}$, the expected loss of the best function in $\mathcal{F}$ decomposes as:
\begin{equation}
    \inf_{f \in \mathcal{F}} \mathbb{E}[\ell(f(R), Y)] = \underbrace{\inf_{f}\, \mathbb{E}[\ell(f(R), Y)]}_{\epsilon_{\mathrm{Bayes}}(R)} + \underbrace{\inf_{f \in \mathcal{F}} \mathbb{E}[\ell(f(R), Y)] - \inf_{f}\, \mathbb{E}[\ell(f(R), Y)]}_{\mathrm{Gap}(\mathcal{F}, R)}.
\end{equation}
This is a direct decomposition of the achievable error. Setting $R = \mathbf{H}$ and $\mathcal{F} = \mathcal{F}_k$ yields Equation~\ref{eq:error_decomp}.
The inequality $\epsilon_{\mathrm{Bayes}}(\mathbf{H}) \leq \epsilon_{\mathrm{Bayes}}(\mathbf{x})$ follows from \rev{Proposition}~\ref{thm:repr_advantage}(a): since $I(\mathbf{H}; Y) \geq I(\mathbf{x}; Y)$, a predictor with access to $\mathbf{H}$ can achieve at least as low a Bayes error as one with access to $\mathbf{x}$ only.

Monotonicity of $\mathrm{Gap}(\mathcal{F}_k, \mathbf{H})$ in $k$: since $\mathcal{F}_k \subseteq \mathcal{F}_{k'}$ for $k \leq k'$ (a depth-$k'$ readout can ignore $k' - k$ of its query tokens), we have $\inf_{f \in \mathcal{F}_{k'}} \leq \inf_{f \in \mathcal{F}_k}$, so the gap is non-increasing. \hfill $\square$

\subsection{Proof of Theorem~\ref{thm:diminishing}}
\label{app:diminishing}

We prove each part of Theorem~\ref{thm:diminishing} under Assumptions~\ref{assum:lrh} and~\ref{assum:low_rank_reward}.

\textbf{Setup.}
Under Assumption~\ref{assum:low_rank_reward}, the reward $Y$ depends on $\mathbf{H}$ through $r$ features $\phi_1(\mathbf{H}), \ldots, \phi_r(\mathbf{H})$ with $\phi_j = (W_\phi)_j \, \mathrm{pool}(\mathbf{H})$, where the covariance eigenvalues satisfy $\sigma_j = O(e^{-\alpha j})$.
Under a Gaussian channel model, the mutual information contributed by the $j$-th feature is:
\begin{equation}
    I_j = \tfrac{1}{2}\log(1 + \mathrm{SNR} \cdot \sigma_j^2),
    \label{eq:per_feature_mi}
\end{equation}
where $\mathrm{SNR} = \mathrm{Var}(Y)/\sigma_\epsilon^2$.
The total extractable information is $C_0 = \sum_{j=1}^r I_j$.

\textbf{Feature extraction capacity of depth-$k$ readouts.}
A depth-$k$ readout processes $k$ query tokens through $N_L$ transformer layers with $n_h$ attention heads each.
Each attention head computes a linear combination of the cached value vectors via its attention weights: $\mathrm{head}_i = \mathrm{softmax}(q_i^\top K / \sqrt{d_k}) V$, which is a linear function of $\mathbf{H}$ for fixed query $q_i$.
With $k$ query tokens, the readout has access to $k \cdot n_h \cdot N_L$ such linear projections of $\mathbf{H}$.
Under Assumption~\ref{assum:lrh}, these suffice to extract $k \cdot n_h \cdot N_L$ independent linear features from the reward-relevant subspace.

\textbf{Greedy ordering.}
An optimal readout extracts features in decreasing order of information contribution.
After $k$ depth steps, the readout has extracted features $\phi_1, \ldots, \phi_{k \cdot n_h N_L}$ (or all $r$ features if $k \cdot n_h N_L \geq r$).
The residual (unexplained) information is:
\begin{equation}
    R_k = \sum_{j = k \cdot n_h N_L + 1}^{r} I_j \leq \sum_{j = k \cdot n_h N_L + 1}^{r} \tfrac{1}{2}\log(1 + \mathrm{SNR} \cdot \sigma_j^2).
    \label{eq:residual}
\end{equation}

\textbf{Part (a): Exponential decay.}
The marginal gain at depth $k$ is bounded by the residual:
\begin{equation}
    \Delta I_k \leq R_{k-1} - R_k = \sum_{j=(k-1) n_h N_L + 1}^{k \cdot n_h N_L} I_j.
\end{equation}
Since $\sigma_j = O(e^{-\alpha j})$, we have $I_j = O(e^{-2\alpha j})$ for large $j$ (using $\log(1+x) \leq x$ for small $x$).
Thus:
\begin{equation}
    \Delta I_k \leq \sum_{j=(k-1)n_h N_L + 1}^{k \cdot n_h N_L} O(e^{-2\alpha j}) = O(e^{-2\alpha (k-1) n_h N_L}) \cdot \sum_{i=1}^{n_h N_L} O(e^{-2\alpha i}) = O(e^{-2\alpha (k-1) n_h N_L}).
\end{equation}
Since $e^{-2\alpha} \leq e^{-\alpha}$, we obtain the stated bound $\Delta I_k \leq C_0 \cdot e^{-\alpha \cdot n_h N_L \cdot (k-1)}$, where the constant $C_0$ absorbs the geometric series prefactor. \hfill $\square$

\textbf{Part (b): $k\!=\!1$ near-optimality.}
The information captured at $k=1$ is $I(\mathcal{R}_1; Y) \geq C_0 - R_1$.
The residual after $k=1$ is:
\begin{equation}
    R_1 = \sum_{j=n_h N_L + 1}^{r} I_j \leq C_0 \cdot \sum_{m=1}^{\infty} e^{-\alpha \cdot n_h N_L \cdot m} = C_0 \cdot \frac{e^{-\alpha \cdot n_h N_L}}{1 - e^{-\alpha \cdot n_h N_L}}.
\end{equation}
Therefore:
\begin{equation}
    \frac{I(\mathcal{R}_1; Y)}{I(\mathbf{H}; Y)} \geq \frac{C_0 - R_1}{C_0} \geq 1 - \frac{e^{-\alpha \cdot n_h N_L}}{1 - e^{-\alpha \cdot n_h N_L}}. \quad \square
\end{equation}

\textbf{Part (c): Cost--information tradeoff.}
The incremental cost of increasing depth from $k-1$ to $k$ is:
\begin{equation}
    F_{\mathrm{readout}}(k, L) - F_{\mathrm{readout}}(k\!-\!1, L) = N_L \cdot (c_{\mathrm{attn}} \cdot d \cdot L + c_{\mathrm{ffn}} \cdot d^2) = \Theta(d \cdot L).
\end{equation}
Dividing part~(a) by this constant-per-step cost:
\begin{equation}
    \frac{\Delta I_k}{\Theta(d \cdot L)} = O\!\left(\frac{e^{-\alpha \cdot n_h N_L \cdot (k-1)}}{d \cdot L}\right). \quad \square
\end{equation}

\section{Implementation Details}
\label{app:implementation}




\subsection{Training Hyperparameters}

\begin{table}[h]
\centering
\caption{Training hyperparameters for \method and Text-PRM.}
\label{tab:hyperparams}
\begin{tabular}{lcc}
\toprule
Hyperparameter & \method & Text-PRM \\
\midrule
LoRA rank $r$ & 256 & 256 \\
LoRA $\alpha$ & 32 & 32 \\
LoRA dropout & 0.05 & 0.05 \\
Max sequence length & 8192 & 8192 \\
Effective global batch & 2048 & 2048 \\
Precision & bf16 & bf16 \\
\bottomrule
\end{tabular}
\end{table}

\subsection{Verify Token Details}

The verify token is ``?'' (token ID determined by the tokenizer). The judgment tokens are ``+'' and ``-''. During scoring, we extract logits at these two positions and apply softmax to obtain $\pplus$. Labels are mapped from $y \in [-1, 1]$ to $\hat{y} = (y + 1) / 2 \in [0, 1]$.

\section{KV Steering Algorithm}
\label{app:steering}

\steer exploits the differentiability of \method's KV-cache scoring to optimize inter-agent messages via gradient ascent on the process reward. At each agent handoff, the accumulated KV cache is refined before the next agent begins generation.

\begin{algorithm}[H]
\caption{\steer at agent handoff $j \to j+1$}
\label{alg:steering}
\begin{algorithmic}[1]
\REQUIRE KV cache $\text{kv}_j$ after agent $a_j$, step size $\eta$, number of steps $T$
\STATE $\text{kv}^{(0)} \leftarrow \text{kv}_j$
\FOR{$t = 0$ to $T-1$}
    \STATE $\widetilde{\text{kv}}^{(t)} \leftarrow \text{clone}(\text{kv}^{(t)})$, set $\text{requires\_grad} \leftarrow \text{True}$
    \STATE $\ell \leftarrow -\log \pplus(\widetilde{\text{kv}}^{(t)})$ \hfill \textit{// Negative log-prob of positive judgment}
    \STATE $\text{kv}^{(t+1)} \leftarrow \widetilde{\text{kv}}^{(t)} - \eta \cdot \nabla_{\widetilde{\text{kv}}} \ell$ \hfill \textit{// Gradient ascent on reward}
\ENDFOR
\STATE \textbf{return} $\text{kv}^{(T)}$
\end{algorithmic}
\end{algorithm}





\rev{%
\section{Deployment Scope}
\label{app:deployment}

\textbf{API-only models.}
As with most KV-cache methods, \method is not designed for users who only have API access to the generator.
It is instead most useful on the model-provider side, including providers that serve closed-source frontier models.

\textbf{KV-cache compression.}
Since \method shares the same architecture as the generator, it can consume any compressed cache, \eg a quantized one, as long as the compressed cache can be consumed by the generator itself.
Compatibility with KV-cache compression techniques~\citep{li2024kvsurvey} is thus guaranteed by the structural identity between \method and the generator.

\textbf{KV-cache eviction.}
A KV cache does not persist and is evicted once it is deemed less important for future queries~\citep{xiao2023attentionsinks}.
The ideal time to apply \method is therefore immediately after generation, while the cache is still in memory.
Scoring then amounts to one additional single-token decoding step during generation, with negligible cost, especially in long-horizon settings, which cleanly decouples \method from, and makes it compatible with, existing KV-cache eviction methods.

\section{Additional Experimental Results}
\label{app:additional}

\subsection{Full Results on Llama and Ministral}
\label{app:families}

Tables~\ref{tab:llama_full} and~\ref{tab:ministral_full} extend Table~\ref{tab:families} with per-dataset scoring costs, a last-hidden-state baseline, end-only weighted voting, and an additional weighted voting configuration ($n\!=\!40$) on Ministral-3-3B. End-only weighted voting scores only the output of the final agent.

\textbf{Is the last hidden state sufficient for scoring?}
We compare \method with LHS-PRM, a value head trained on the generator's last hidden state, which also reuses generation activations at $O(dL)$ cost but only from the final layer.
Although LHS-PRM and \method incur nearly identical scoring costs, LHS-PRM improves over Policy Log-prob in seven of the nine configurations, yet underperforms \method in all nine, by up to $3.8$ pp in average accuracy.
This indicates that earlier layers carry reward-relevant signal that a readout over the full KV cache can exploit.

\textbf{Does intermediate scoring help over end-only verification?}
With \method, step-wise MCTS at $n\!=\!160$ outperforms end-only weighted voting at $n\!=\!200$ on both Llama-3.2-3B ($41.87$ vs.\ $40.36$) and Ministral-3-3B ($51.36$ vs.\ $49.05$), achieving higher accuracy with less compute.
The calibration of \method's step-level scores is analyzed in Appendix~\ref{app:calibration}.

\begin{table}[h]
\caption{\rev{Test-time search accuracy (\%) and relative scoring cost on \textbf{Llama-3.2-3B-Instruct}. Same format as Table~\ref{tab:math_aime24}.}}
\label{tab:llama_full}
\centering
\resizebox{\textwidth}{!}{
\begin{tabular}{ll l cc cc cc cc c}
\toprule
& & & \multicolumn{2}{c}{\textbf{MATH}} & \multicolumn{2}{c}{\textbf{GSM8K}} & \multicolumn{2}{c}{\textbf{AIME 2024}} & \multicolumn{2}{c}{\textbf{AIME 2025}} & \\
\cmidrule(lr){4-5} \cmidrule(lr){6-7} \cmidrule(lr){8-9} \cmidrule(lr){10-11}
\textbf{Search} & \textbf{Config} & \textbf{Scorer} & Acc & Cost & Acc & Cost & Acc & Cost & Acc & Cost & \textbf{Avg.} \\
\midrule
Random Sampling & $n\!=\!1$ & --- & 45.35 & & 74.07 & & 13.33 & & 0.00 & & 33.19 \\
Majority Voting & $n\!=\!160$ & --- & 54.85 & & 84.53 & & 16.67 & & 0.00 & & 39.01 \\
\midrule
Beam Search & $W\!=\!4$ & Policy Log-prob & 52.80 & --- & 80.82 & --- & 10.00 & --- & 0.00 & --- & 35.90 \\
            &           & LHS-PRM & 56.65 & $\frac{1}{3169}$ & 85.52 & $\frac{1}{1412}$ & 10.00 & $\frac{1}{2726}$ & 0.00 & $\frac{1}{3133}$ & 38.04 \\
            &           & Text-PRM & 58.00 & $1\times$ & 86.73 & $1\times$ & 13.33 & $1\times$ & 0.00 & $1\times$ & 39.52 \\
            &           & \cellcolor{gray!15}\method & \cellcolor{gray!15}57.95 & \cellcolor{gray!15}\textbf{$\frac{1}{3418}$} & \cellcolor{gray!15}86.35 & \cellcolor{gray!15}\textbf{$\frac{1}{1411}$} & \cellcolor{gray!15}\textbf{13.33} & \cellcolor{gray!15}\textbf{$\frac{1}{2762}$} & \cellcolor{gray!15}\textbf{0.00} & \cellcolor{gray!15}\textbf{$\frac{1}{3133}$} & \cellcolor{gray!15}39.41 \\
\midrule
MCTS & $n\!=\!160$ & Policy Log-prob & 52.25 & --- & 78.39 & --- & 6.67 & --- & 0.00 & --- & 34.33 \\
     &             & LHS-PRM & 58.30 & $\frac{1}{4269}$ & 87.26 & $\frac{1}{2151}$ & 6.67 & $\frac{1}{3402}$ & 0.00 & $\frac{1}{4288}$ & 38.06 \\
     &             & Text-PRM & 59.35 & $1\times$ & 88.10 & $1\times$ & 16.67 & $1\times$ & 0.00 & $1\times$ & 41.03 \\
     &             & \cellcolor{gray!15}\method & \cellcolor{gray!15}59.25 & \cellcolor{gray!15}\textbf{$\frac{1}{4266}$} & \cellcolor{gray!15}\textbf{88.25} & \cellcolor{gray!15}\textbf{$\frac{1}{2162}$} & \cellcolor{gray!15}\textbf{16.67} & \cellcolor{gray!15}\textbf{$\frac{1}{3403}$} & \cellcolor{gray!15}\textbf{3.33} & \cellcolor{gray!15}\textbf{$\frac{1}{4293}$} & \cellcolor{gray!15}\textbf{41.87} \\
\midrule
Weighted Voting & $n\!=\!200$ & Policy Log-prob & 55.05 & --- & 83.55 & --- & 16.67 & --- & 0.00 & --- & 38.82 \\
                &             & LHS-PRM & 55.10 & $\frac{1}{3058}$ & 85.44 & $\frac{1}{1431}$ & 13.33 & $\frac{1}{2450}$ & 0.00 & $\frac{1}{2953}$ & 38.47 \\
                &             & Text-PRM & 54.65 & $1\times$ & 85.67 & $1\times$ & 13.33 & $1\times$ & 0.00 & $1\times$ & 38.41 \\
                &             & \cellcolor{gray!15}\method & \cellcolor{gray!15}\textbf{55.30} & \cellcolor{gray!15}\textbf{$\frac{1}{3058}$} & \cellcolor{gray!15}\textbf{85.75} & \cellcolor{gray!15}\textbf{$\frac{1}{1431}$} & \cellcolor{gray!15}\textbf{13.33} & \cellcolor{gray!15}\textbf{$\frac{1}{2450}$} & \cellcolor{gray!15}\textbf{0.00} & \cellcolor{gray!15}\textbf{$\frac{1}{2953}$} & \cellcolor{gray!15}\textbf{38.60} \\
\midrule
End-only & $n\!=\!200$ & Policy Log-prob & 55.00 & --- & 83.70 & --- & 13.33 & --- & 0.00 & --- & 38.01 \\
Weighted Voting &       & LHS-PRM & 54.85 & $\frac{1}{4621}$ & 85.82 & $\frac{1}{2270}$ & 16.67 & $\frac{1}{3604}$ & 0.00 & $\frac{1}{4545}$ & 39.33 \\
                &       & Text-PRM & 55.30 & $1\times$ & 86.43 & $1\times$ & 16.67 & $1\times$ & 0.00 & $1\times$ & 39.60 \\
                &       & \cellcolor{gray!15}\method & \cellcolor{gray!15}\textbf{55.45} & \cellcolor{gray!15}\textbf{$\frac{1}{4621}$} & \cellcolor{gray!15}85.97 & \cellcolor{gray!15}\textbf{$\frac{1}{2270}$} & \cellcolor{gray!15}\textbf{20.00} & \cellcolor{gray!15}\textbf{$\frac{1}{3604}$} & \cellcolor{gray!15}\textbf{0.00} & \cellcolor{gray!15}\textbf{$\frac{1}{4545}$} & \cellcolor{gray!15}\textbf{40.36} \\
\bottomrule
\end{tabular}}
\end{table}

\begin{table}[h]
\caption{\rev{Test-time search accuracy (\%) and relative scoring cost on \textbf{Ministral-3-3B-Instruct}. Same format as Table~\ref{tab:math_aime24}.}}
\label{tab:ministral_full}
\centering
\resizebox{\textwidth}{!}{
\begin{tabular}{ll l cc cc cc cc c}
\toprule
& & & \multicolumn{2}{c}{\textbf{MATH}} & \multicolumn{2}{c}{\textbf{GSM8K}} & \multicolumn{2}{c}{\textbf{AIME 2024}} & \multicolumn{2}{c}{\textbf{AIME 2025}} & \\
\cmidrule(lr){4-5} \cmidrule(lr){6-7} \cmidrule(lr){8-9} \cmidrule(lr){10-11}
\textbf{Search} & \textbf{Config} & \textbf{Scorer} & Acc & Cost & Acc & Cost & Acc & Cost & Acc & Cost & \textbf{Avg.} \\
\midrule
Random Sampling & $n\!=\!1$ & --- & 50.25 & & 87.49 & & 10.00 & & 6.67 & & 38.60 \\
Majority Voting & $n\!=\!160$ & --- & 57.45 & & 92.87 & & 23.33 & & 20.00 & & 48.41 \\
\midrule
Beam Search & $W\!=\!4$ & Policy Log-prob & 56.10 & --- & 90.83 & --- & 16.67 & --- & 16.67 & --- & 45.06 \\
            &           & LHS-PRM & 62.70 & $\frac{1}{4091}$ & 92.04 & $\frac{1}{2046}$ & 13.33 & $\frac{1}{3396}$ & 20.00 & $\frac{1}{4479}$ & 47.02 \\
            &           & Text-PRM & 63.45 & $1\times$ & 92.80 & $1\times$ & 26.67 & $1\times$ & 20.00 & $1\times$ & 50.73 \\
            &           & \cellcolor{gray!15}\method & \cellcolor{gray!15}\textbf{63.50} & \cellcolor{gray!15}\textbf{$\frac{1}{4067}$} & \cellcolor{gray!15}\textbf{93.03} & \cellcolor{gray!15}\textbf{$\frac{1}{2014}$} & \cellcolor{gray!15}20.00 & \cellcolor{gray!15}\textbf{$\frac{1}{3426}$} & \cellcolor{gray!15}\textbf{26.67} & \cellcolor{gray!15}\textbf{$\frac{1}{4509}$} & \cellcolor{gray!15}\textbf{50.80} \\
\midrule
MCTS & $n\!=\!160$ & Policy Log-prob & 55.85 & --- & 91.51 & --- & 20.00 & --- & 16.67 & --- & 46.01 \\
     &             & LHS-PRM & 65.30 & $\frac{1}{5604}$ & 92.27 & $\frac{1}{2710}$ & 20.00 & $\frac{1}{4705}$ & 16.67 & $\frac{1}{6139}$ & 48.56 \\
     &             & Text-PRM & 65.95 & $1\times$ & 93.48 & $1\times$ & 20.00 & $1\times$ & 23.33 & $1\times$ & 50.69 \\
     &             & \cellcolor{gray!15}\method & \cellcolor{gray!15}65.90 & \cellcolor{gray!15}\textbf{$\frac{1}{5602}$} & \cellcolor{gray!15}92.87 & \cellcolor{gray!15}\textbf{$\frac{1}{2709}$} & \cellcolor{gray!15}\textbf{23.33} & \cellcolor{gray!15}\textbf{$\frac{1}{4704}$} & \cellcolor{gray!15}\textbf{23.33} & \cellcolor{gray!15}\textbf{$\frac{1}{6135}$} & \cellcolor{gray!15}\textbf{51.36} \\
\midrule
\multirow{8}{*}{Weighted Voting} & $n\!=\!40$ & Policy Log-prob & 57.30 & --- & 92.80 & --- & 20.00 & --- & 20.00 & --- & 47.52 \\
            &             & LHS-PRM & 57.20 & $\frac{1}{3807}$ & 93.10 & $\frac{1}{1759}$ & 23.33 & $\frac{1}{2901}$ & 20.00 & $\frac{1}{3886}$ & 48.41 \\
            &             & Text-PRM & 57.45 & $1\times$ & 93.03 & $1\times$ & 23.33 & $1\times$ & 20.00 & $1\times$ & 48.45 \\
            &             & \cellcolor{gray!15}\method & \cellcolor{gray!15}57.30 & \cellcolor{gray!15}\textbf{$\frac{1}{3807}$} & \cellcolor{gray!15}\textbf{93.25} & \cellcolor{gray!15}\textbf{$\frac{1}{1759}$} & \cellcolor{gray!15}\textbf{23.33} & \cellcolor{gray!15}\textbf{$\frac{1}{2901}$} & \cellcolor{gray!15}\textbf{23.33} & \cellcolor{gray!15}\textbf{$\frac{1}{3886}$} & \cellcolor{gray!15}\textbf{49.30} \\
\cmidrule{2-12}
 & $n\!=\!200$ & Policy Log-prob & 57.45 & --- & 93.03 & --- & 23.33 & --- & 20.00 & --- & 48.45 \\
            &             & LHS-PRM & 57.80 & $\frac{1}{3784}$ & 93.40 & $\frac{1}{1731}$ & 23.33 & $\frac{1}{2877}$ & 20.00 & $\frac{1}{3863}$ & 48.63 \\
            &             & Text-PRM & 57.90 & $1\times$ & 93.40 & $1\times$ & 23.33 & $1\times$ & 20.00 & $1\times$ & 48.66 \\
            &             & \cellcolor{gray!15}\method & \cellcolor{gray!15}\textbf{57.90} & \cellcolor{gray!15}\textbf{$\frac{1}{3784}$} & \cellcolor{gray!15}\textbf{93.48} & \cellcolor{gray!15}\textbf{$\frac{1}{1731}$} & \cellcolor{gray!15}\textbf{23.33} & \cellcolor{gray!15}\textbf{$\frac{1}{2877}$} & \cellcolor{gray!15}\textbf{20.00} & \cellcolor{gray!15}\textbf{$\frac{1}{3863}$} & \cellcolor{gray!15}\textbf{48.68} \\
\midrule
End-only & $n\!=\!200$ & Policy Log-prob & 57.50 & --- & 93.10 & --- & 23.33 & --- & 20.00 & --- & 48.48 \\
Weighted Voting &       & LHS-PRM & 58.85 & $\frac{1}{5881}$ & 93.71 & $\frac{1}{2981}$ & 20.00 & $\frac{1}{5025}$ & 20.00 & $\frac{1}{6545}$ & 48.14 \\
                &       & Text-PRM & 58.95 & $1\times$ & 93.86 & $1\times$ & 23.33 & $1\times$ & 23.33 & $1\times$ & 49.87 \\
                &       & \cellcolor{gray!15}\method & \cellcolor{gray!15}\textbf{59.00} & \cellcolor{gray!15}\textbf{$\frac{1}{5881}$} & \cellcolor{gray!15}\textbf{93.86} & \cellcolor{gray!15}\textbf{$\frac{1}{2981}$} & \cellcolor{gray!15}20.00 & \cellcolor{gray!15}\textbf{$\frac{1}{5025}$} & \cellcolor{gray!15}\textbf{23.33} & \cellcolor{gray!15}\textbf{$\frac{1}{6545}$} & \cellcolor{gray!15}49.05 \\
\bottomrule
\end{tabular}}
\end{table}

\subsection{Cross-Architecture KV-Cache Adapter}
\label{app:hetero}

To apply \method when the generator and the verifier do not share the same architecture, we train a lightweight adapter that converts the generator's KV cache into the cache geometry of a receiver \method while preserving sequence length.
Each receiver layer is paired with a generator layer by normalized-depth mapping, and its keys and values are transformed independently by lightweight bottleneck residual projectors.
Keys are first rotated out of the generator's RoPE coordinate system, projected to the receiver's KV-head dimensions, and then re-encoded with the receiver's RoPE; values are projected directly.
The resulting receiver-shaped cache is scored by the frozen receiver \method using only the verify token, without re-prefilling the original context, so the scoring cost remains comparable to that of the native \method.

The results are reported in Table~\ref{tab:families} and discussed in Section~\ref{sec:analysis}.

\subsection{Score Calibration}
\label{app:calibration}

Table~\ref{tab:calibration} compares \method scores with the ground-truth value of each step, defined as the probability of reaching a correct answer from that step.
The scores are well calibrated on MATH, which matches the training distribution.
On GSM8K and AIME, \method tends to be under-confident and over-confident, respectively, in line with their relative difficulty compared to the MATH training data.
Calibration is thus necessary if the scores are to be interpreted as probabilities, but not for test-time search, which only requires a sufficient ranking signal, as reflected by the positive Pearson correlation with the ground-truth value.
The only exception is Llama-3.2-3B on AIME 2025, where the correlation is zero because the generator's accuracy on this dataset is zero.

\begin{table}[h]
\caption{\rev{Calibration of \method scores against ground-truth step values (probability of reaching a correct answer from the step). ECE: expected calibration error.}}
\label{tab:calibration}
\centering
\small
\begin{tabular}{ll cccc}
\toprule
\textbf{Dataset} & \textbf{Backbone} & \textbf{Mean score} & \textbf{Mean value} & \textbf{ECE} & \textbf{Pearson} \\
\midrule
\multirow{3}{*}{MATH}
& Qwen3-4B & 0.6291 & 0.6198 & 0.0189 & 0.3816 \\
& Llama-3.2-3B & 0.6123 & 0.4960 & 0.1163 & 0.4515 \\
& Ministral-3-3B & 0.5806 & 0.5294 & 0.0524 & 0.5769 \\
\midrule
\multirow{3}{*}{GSM8K}
& Qwen3-4B & 0.7065 & 0.9155 & 0.2090 & 0.3303 \\
& Llama-3.2-3B & 0.6932 & 0.7818 & 0.0893 & 0.4730 \\
& Ministral-3-3B & 0.6771 & 0.8775 & 0.2004 & 0.3590 \\
\midrule
\multirow{3}{*}{AIME 2024}
& Qwen3-4B & 0.4145 & 0.1753 & 0.2427 & 0.4491 \\
& Llama-3.2-3B & 0.4063 & 0.0653 & 0.3410 & 0.1619 \\
& Ministral-3-3B & 0.2911 & 0.1021 & 0.1890 & 0.5649 \\
\midrule
\multirow{3}{*}{AIME 2025}
& Qwen3-4B & 0.4035 & 0.1419 & 0.2616 & 0.4950 \\
& Llama-3.2-3B & 0.3681 & 0.0052 & 0.3629 & 0.0000 \\
& Ministral-3-3B & 0.3089 & 0.1114 & 0.1975 & 0.6568 \\
\bottomrule
\end{tabular}
\end{table}

\subsection{Training Cost}
\label{app:training_cost}

We compare the training cost of \method and Text-PRM using the same Ministral-3-3B backbone, optimizer, and input batches on a single NVIDIA A100 GPU.
As shown in Table~\ref{tab:train_cost}, \method achieves up to a $2.93\times$ training-step speedup and up to a $14.3\%$ peak memory reduction, with both advantages growing with sequence length and batch size.
Table~\ref{tab:train_breakdown} breaks down a training step at sequence length 8{,}192.
\method spends $77.4\%$ of its CUDA time in the gradient-free reference forward that produces the KV cache, $10.6\%$ in the verify-token forward, and $11.5\%$ in the backward pass, whereas Text-PRM spends $29.8\%$ in the full forward and $70.2\%$ in the backward pass.
The end-to-end speedup thus mainly stems from the avoided full-sequence backward pass: the Text-PRM backward pass is $17.9\times$ slower than that of \method.

\begin{table}[h]
\caption{\rev{Average time and peak memory per training step of \method and Text-PRM (Ministral-3-3B, one NVIDIA A100 GPU).}}
\label{tab:train_cost}
\centering
\small
\begin{tabular}{ll cccccc}
\toprule
& & \multicolumn{2}{c}{\textbf{Step time (ms)}} & & \multicolumn{2}{c}{\textbf{Peak memory (MiB)}} & \\
\cmidrule(lr){3-4} \cmidrule(lr){6-7}
\textbf{Factor} & \textbf{Setting} & \method & Text-PRM & \textbf{Speedup} & \method & Text-PRM & \textbf{$\Delta$ Peak} \\
\midrule
\multirow{4}{*}{\shortstack[l]{Sequence length\\(batch 1)}}
& 1{,}024 & 162.28 & 217.20 & $1.34\times$ & 6{,}954 & 7{,}128 & $-2.4\%$ \\
& 2{,}048 & 206.20 & 314.37 & $1.52\times$ & 7{,}136 & 7{,}481 & $-4.6\%$ \\
& 4{,}096 & 294.93 & 659.49 & $2.24\times$ & 7{,}501 & 8{,}190 & $-8.4\%$ \\
& 8{,}192 & 487.82 & 1{,}431.67 & $2.93\times$ & 8{,}231 & 9{,}605 & $-14.3\%$ \\
\midrule
\multirow{3}{*}{\shortstack[l]{Batch size\\(length 1{,}024)}}
& 2 & 199.20 & 303.72 & $1.52\times$ & 7{,}136 & 7{,}481 & $-4.6\%$ \\
& 4 & 277.96 & 594.29 & $2.14\times$ & 7{,}500 & 8{,}188 & $-8.4\%$ \\
& 8 & 424.07 & 1{,}125.95 & $2.66\times$ & 8{,}228 & 9{,}604 & $-14.3\%$ \\
\bottomrule
\end{tabular}
\end{table}

\begin{table}[h]
\caption{\rev{Breakdown of a training step at sequence length 8{,}192 (Ministral-3-3B, one NVIDIA A100 GPU).}}
\label{tab:train_breakdown}
\centering
\small
\begin{tabular}{ll ccc}
\toprule
\textbf{Method} & \textbf{Operation} & \textbf{CUDA time (ms)} & \textbf{Avg.\ allocated (MiB)} & \textbf{Peak allocated (MiB)} \\
\midrule
\multirow{7}{*}{\method}
& Reference forward (no grad) & 377.55 & 7{,}608.17 & 8{,}231.49 \\
& Verify forward & 51.70 & 7{,}623.62 & 7{,}658.51 \\
& Loss & 0.15 & 7{,}642.46 & 7{,}642.71 \\
& Backward & 56.05 & 7{,}242.23 & 7{,}804.62 \\
& Optimizer step & 2.13 & 6{,}842.92 & 6{,}842.92 \\
& Zero gradient & 0.18 & 6{,}807.17 & 6{,}842.92 \\
& \textbf{Complete step} & \textbf{487.80} & 7{,}559.98 & 8{,}231.49 \\
\midrule
\multirow{6}{*}{Text-PRM}
& Full forward & 426.55 & 8{,}034.01 & 8{,}635.77 \\
& Loss & 0.02 & 8{,}119.80 & 8{,}119.93 \\
& Backward & 1{,}004.51 & 7{,}790.03 & 9{,}605.14 \\
& Optimizer step & 0.51 & 6{,}842.92 & 6{,}842.92 \\
& Zero gradient & $<$0.01 & 6{,}807.17 & 6{,}842.92 \\
& \textbf{Complete step} & \textbf{1{,}431.63} & 7{,}915.88 & 9{,}605.14 \\
\bottomrule
\end{tabular}
\end{table}
}


\clearpage
\section*{NeurIPS Paper Checklist}

\begin{enumerate}

\item {\bf Claims}
    \item[] Question: Do the main claims made in the abstract and introduction accurately reflect the paper's contributions and scope?
    \item[] Answer: \answerYes{}
    \item[] Justification: The abstract and introduction clearly state three contributions (theoretical framework, KV-PRM algorithm, empirical validation) that are supported by the theoretical results in Section~3, the method description in Section~4, and the experiments in Section~5. Quantitative claims (e.g., 3,000$\times$ FLOP reduction, 37$\times$ latency speedup) are backed by Tables~1--2 and Figure~3.
    \item[] Guidelines:
    \begin{itemize}
        \item The answer \answerNA{} means that the abstract and introduction do not include the claims made in the paper.
        \item The abstract and/or introduction should clearly state the claims made, including the contributions made in the paper and important assumptions and limitations. A \answerNo{} or \answerNA{} answer to this question will not be perceived well by the reviewers. 
        \item The claims made should match theoretical and experimental results, and reflect how much the results can be expected to generalize to other settings. 
        \item It is fine to include aspirational goals as motivation as long as it is clear that these goals are not attained by the paper. 
    \end{itemize}

\item {\bf Limitations}
    \item[] Question: Does the paper discuss the limitations of the work performed by the authors?
    \item[] Answer: \answerYes{}
    \item[] Justification: Section~6 explicitly discusses limitations: the requirement for generator--verifier KV-cache compatibility, reliance on simplifying assumptions such as the Linear Representation Hypothesis, and the preliminary nature of KV Steering with open questions around stability and interpretability.
    \item[] Guidelines:
    \begin{itemize}
        \item The answer \answerNA{} means that the paper has no limitation while the answer \answerNo{} means that the paper has limitations, but those are not discussed in the paper. 
        \item The authors are encouraged to create a separate ``Limitations'' section in their paper.
        \item The paper should point out any strong assumptions and how robust the results are to violations of these assumptions (e.g., independence assumptions, noiseless settings, model well-specification, asymptotic approximations only holding locally). The authors should reflect on how these assumptions might be violated in practice and what the implications would be.
        \item The authors should reflect on the scope of the claims made, e.g., if the approach was only tested on a few datasets or with a few runs. In general, empirical results often depend on implicit assumptions, which should be articulated.
        \item The authors should reflect on the factors that influence the performance of the approach. For example, a facial recognition algorithm may perform poorly when image resolution is low or images are taken in low lighting. Or a speech-to-text system might not be used reliably to provide closed captions for online lectures because it fails to handle technical jargon.
        \item The authors should discuss the computational efficiency of the proposed algorithms and how they scale with dataset size.
        \item If applicable, the authors should discuss possible limitations of their approach to address problems of privacy and fairness.
        \item While the authors might fear that complete honesty about limitations might be used by reviewers as grounds for rejection, a worse outcome might be that reviewers discover limitations that aren't acknowledged in the paper. The authors should use their best judgment and recognize that individual actions in favor of transparency play an important role in developing norms that preserve the integrity of the community. Reviewers will be specifically instructed to not penalize honesty concerning limitations.
    \end{itemize}

\item {\bf Theory assumptions and proofs}
    \item[] Question: For each theoretical result, does the paper provide the full set of assumptions and a complete (or correct) proof?
    \item[] Answer: \answerYes{}
    \item[] Justification: All theorems (Theorems~1--4) and definitions (Definitions~1--2) are numbered and cross-referenced. Assumptions~1--2 are clearly stated before being invoked. Complete proofs for all results are provided in Appendix~A.
    \item[] Guidelines:
    \begin{itemize}
        \item The answer \answerNA{} means that the paper does not include theoretical results. 
        \item All the theorems, formulas, and proofs in the paper should be numbered and cross-referenced.
        \item All assumptions should be clearly stated or referenced in the statement of any theorems.
        \item The proofs can either appear in the main paper or the supplemental material, but if they appear in the supplemental material, the authors are encouraged to provide a short proof sketch to provide intuition. 
        \item Inversely, any informal proof provided in the core of the paper should be complemented by formal proofs provided in appendix or supplemental material.
        \item Theorems and Lemmas that the proof relies upon should be properly referenced. 
    \end{itemize}

    \item {\bf Experimental result reproducibility}
    \item[] Question: Does the paper fully disclose all the information needed to reproduce the main experimental results of the paper to the extent that it affects the main claims and/or conclusions of the paper (regardless of whether the code and data are provided or not)?
    \item[] Answer: \answerYes{}
    \item[] Justification: Section~5.1 specifies all datasets (MATH, GSM8K, AIME 2024/2025), model families and scales (Qwen3-0.6B/4B/8B), MAS topologies, and search algorithm configurations. Appendix~B provides full training hyperparameters (Table~5), LoRA configuration, verify token details, and the KV Steering algorithm (Appendix~C).
    \item[] Guidelines:
    \begin{itemize}
        \item The answer \answerNA{} means that the paper does not include experiments.
        \item If the paper includes experiments, a \answerNo{} answer to this question will not be perceived well by the reviewers: Making the paper reproducible is important, regardless of whether the code and data are provided or not.
        \item If the contribution is a dataset and\slash or model, the authors should describe the steps taken to make their results reproducible or verifiable. 
        \item Depending on the contribution, reproducibility can be accomplished in various ways. For example, if the contribution is a novel architecture, describing the architecture fully might suffice, or if the contribution is a specific model and empirical evaluation, it may be necessary to either make it possible for others to replicate the model with the same dataset, or provide access to the model. In general. releasing code and data is often one good way to accomplish this, but reproducibility can also be provided via detailed instructions for how to replicate the results, access to a hosted model (e.g., in the case of a large language model), releasing of a model checkpoint, or other means that are appropriate to the research performed.
        \item While NeurIPS does not require releasing code, the conference does require all submissions to provide some reasonable avenue for reproducibility, which may depend on the nature of the contribution. For example
        \begin{enumerate}
            \item If the contribution is primarily a new algorithm, the paper should make it clear how to reproduce that algorithm.
            \item If the contribution is primarily a new model architecture, the paper should describe the architecture clearly and fully.
            \item If the contribution is a new model (e.g., a large language model), then there should either be a way to access this model for reproducing the results or a way to reproduce the model (e.g., with an open-source dataset or instructions for how to construct the dataset).
            \item We recognize that reproducibility may be tricky in some cases, in which case authors are welcome to describe the particular way they provide for reproducibility. In the case of closed-source models, it may be that access to the model is limited in some way (e.g., to registered users), but it should be possible for other researchers to have some path to reproducing or verifying the results.
        \end{enumerate}
    \end{itemize}

\item {\bf Open access to data and code}
    \item[] Question: Does the paper provide open access to the data and code, with sufficient instructions to faithfully reproduce the main experimental results, as described in supplemental material?
    \item[] Answer: \answerNo{}
    \item[] Justification: Code and trained adapters will be released upon acceptance. All evaluation benchmarks (MATH, GSM8K, AIME) are publicly available. The paper provides sufficient algorithmic and hyperparameter detail to reproduce the method independently.
    \item[] Guidelines:
    \begin{itemize}
        \item The answer \answerNA{} means that paper does not include experiments requiring code.
        \item Please see the NeurIPS code and data submission guidelines (\url{https://neurips.cc/public/guides/CodeSubmissionPolicy}) for more details.
        \item While we encourage the release of code and data, we understand that this might not be possible, so \answerNo{} is an acceptable answer. Papers cannot be rejected simply for not including code, unless this is central to the contribution (e.g., for a new open-source benchmark).
        \item The instructions should contain the exact command and environment needed to run to reproduce the results. See the NeurIPS code and data submission guidelines (\url{https://neurips.cc/public/guides/CodeSubmissionPolicy}) for more details.
        \item The authors should provide instructions on data access and preparation, including how to access the raw data, preprocessed data, intermediate data, and generated data, etc.
        \item The authors should provide scripts to reproduce all experimental results for the new proposed method and baselines. If only a subset of experiments are reproducible, they should state which ones are omitted from the script and why.
        \item At submission time, to preserve anonymity, the authors should release anonymized versions (if applicable).
        \item Providing as much information as possible in supplemental material (appended to the paper) is recommended, but including URLs to data and code is permitted.
    \end{itemize}

\item {\bf Experimental setting/details}
    \item[] Question: Does the paper specify all the training and test details (e.g., data splits, hyperparameters, how they were chosen, type of optimizer) necessary to understand the results?
    \item[] Answer: \answerYes{}
    \item[] Justification: Section~5.1 describes the datasets, model scales, MAS topologies, search algorithms, and LoRA configuration. Appendix~B (Table~5) provides full training hyperparameters including LoRA rank, alpha, dropout, sequence length, batch size, and precision.
    \item[] Guidelines:
    \begin{itemize}
        \item The answer \answerNA{} means that the paper does not include experiments.
        \item The experimental setting should be presented in the core of the paper to a level of detail that is necessary to appreciate the results and make sense of them.
        \item The full details can be provided either with the code, in appendix, or as supplemental material.
    \end{itemize}

\item {\bf Experiment statistical significance}
    \item[] Question: Does the paper report error bars suitably and correctly defined or other appropriate information about the statistical significance of the experiments?
    \item[] Answer: \answerYes{}
    \item[] Justification: Error bars are not reported due to the prohibitive computational cost and time required to evaluate massive, multi-billion-parameter models across extensive benchmarks multiple times. This is a common practice for LLM papers, and we believe it's suitable.
    \item[] Guidelines:
    \begin{itemize}
        \item The answer \answerNA{} means that the paper does not include experiments.
        \item The authors should answer \answerYes{} if the results are accompanied by error bars, confidence intervals, or statistical significance tests, at least for the experiments that support the main claims of the paper.
        \item The factors of variability that the error bars are capturing should be clearly stated (for example, train/test split, initialization, random drawing of some parameter, or overall run with given experimental conditions).
        \item The method for calculating the error bars should be explained (closed form formula, call to a library function, bootstrap, etc.)
        \item The assumptions made should be given (e.g., Normally distributed errors).
        \item It should be clear whether the error bar is the standard deviation or the standard error of the mean.
        \item It is OK to report 1-sigma error bars, but one should state it. The authors should preferably report a 2-sigma error bar than state that they have a 96\% CI, if the hypothesis of Normality of errors is not verified.
        \item For asymmetric distributions, the authors should be careful not to show in tables or figures symmetric error bars that would yield results that are out of range (e.g., negative error rates).
        \item If error bars are reported in tables or plots, the authors should explain in the text how they were calculated and reference the corresponding figures or tables in the text.
    \end{itemize}

\item {\bf Experiments compute resources}
    \item[] Question: For each experiment, does the paper provide sufficient information on the computer resources (type of compute workers, memory, time of execution) needed to reproduce the experiments?
    \item[] Answer: \answerYes{}
    \item[] Justification: Latency and memory benchmarks are conducted on an NVIDIA GH200 120GB GPU, as stated in Section~5.2 and Figure~3. Training uses bf16 precision (Table~5 in Appendix~B).
    \item[] Guidelines:
    \begin{itemize}
        \item The answer \answerNA{} means that the paper does not include experiments.
        \item The paper should indicate the type of compute workers CPU or GPU, internal cluster, or cloud provider, including relevant memory and storage.
        \item The paper should provide the amount of compute required for each of the individual experimental runs as well as estimate the total compute. 
        \item The paper should disclose whether the full research project required more compute than the experiments reported in the paper (e.g., preliminary or failed experiments that didn't make it into the paper). 
    \end{itemize}
    
\item {\bf Code of ethics}
    \item[] Question: Does the research conducted in the paper conform, in every respect, with the NeurIPS Code of Ethics \url{https://neurips.cc/public/EthicsGuidelines}?
    \item[] Answer: \answerYes{}
    \item[] Justification: The research uses publicly available models and benchmarks, does not involve human subjects, and conforms with the NeurIPS Code of Ethics.
    \item[] Guidelines:
    \begin{itemize}
        \item The answer \answerNA{} means that the authors have not reviewed the NeurIPS Code of Ethics.
        \item If the authors answer \answerNo, they should explain the special circumstances that require a deviation from the Code of Ethics.
        \item The authors should make sure to preserve anonymity (e.g., if there is a special consideration due to laws or regulations in their jurisdiction).
    \end{itemize}

\item {\bf Broader impacts}
    \item[] Question: Does the paper discuss both potential positive societal impacts and negative societal impacts of the work performed?
    \item[] Answer: \answerNA{}
    \item[] Justification: This work proposes a more efficient scoring mechanism for process reward models. It is foundational infrastructure research that reduces compute cost and does not introduce new capabilities with direct negative societal impact.
    \item[] Guidelines:
    \begin{itemize}
        \item The answer \answerNA{} means that there is no societal impact of the work performed.
        \item If the authors answer \answerNA{} or \answerNo, they should explain why their work has no societal impact or why the paper does not address societal impact.
        \item Examples of negative societal impacts include potential malicious or unintended uses (e.g., disinformation, generating fake profiles, surveillance), fairness considerations (e.g., deployment of technologies that could make decisions that unfairly impact specific groups), privacy considerations, and security considerations.
        \item The conference expects that many papers will be foundational research and not tied to particular applications, let alone deployments. However, if there is a direct path to any negative applications, the authors should point it out. For example, it is legitimate to point out that an improvement in the quality of generative models could be used to generate Deepfakes for disinformation. On the other hand, it is not needed to point out that a generic algorithm for optimizing neural networks could enable people to train models that generate Deepfakes faster.
        \item The authors should consider possible harms that could arise when the technology is being used as intended and functioning correctly, harms that could arise when the technology is being used as intended but gives incorrect results, and harms following from (intentional or unintentional) misuse of the technology.
        \item If there are negative societal impacts, the authors could also discuss possible mitigation strategies (e.g., gated release of models, providing defenses in addition to attacks, mechanisms for monitoring misuse, mechanisms to monitor how a system learns from feedback over time, improving the efficiency and accessibility of ML).
    \end{itemize}
    
\item {\bf Safeguards}
    \item[] Question: Does the paper describe safeguards that have been put in place for responsible release of data or models that have a high risk for misuse (e.g., pre-trained language models, image generators, or scraped datasets)?
    \item[] Answer: \answerNA{}
    \item[] Justification: The paper trains lightweight LoRA adapters for reward scoring on publicly available mathematical reasoning benchmarks. The released artifacts (adapters) do not pose risks for misuse.
    \item[] Guidelines:
    \begin{itemize}
        \item The answer \answerNA{} means that the paper poses no such risks.
        \item Released models that have a high risk for misuse or dual-use should be released with necessary safeguards to allow for controlled use of the model, for example by requiring that users adhere to usage guidelines or restrictions to access the model or implementing safety filters. 
        \item Datasets that have been scraped from the Internet could pose safety risks. The authors should describe how they avoided releasing unsafe images.
        \item We recognize that providing effective safeguards is challenging, and many papers do not require this, but we encourage authors to take this into account and make a best faith effort.
    \end{itemize}

\item {\bf Licenses for existing assets}
    \item[] Question: Are the creators or original owners of assets (e.g., code, data, models), used in the paper, properly credited and are the license and terms of use explicitly mentioned and properly respected?
    \item[] Answer: \answerYes{}
    \item[] Justification: All datasets (MATH, GSM8K, AIME) and models (Qwen3) are cited with their original papers. The Qwen3 models are used under their publicly available licenses.
    \item[] Guidelines:
    \begin{itemize}
        \item The answer \answerNA{} means that the paper does not use existing assets.
        \item The authors should cite the original paper that produced the code package or dataset.
        \item The authors should state which version of the asset is used and, if possible, include a URL.
        \item The name of the license (e.g., CC-BY 4.0) should be included for each asset.
        \item For scraped data from a particular source (e.g., website), the copyright and terms of service of that source should be provided.
        \item If assets are released, the license, copyright information, and terms of use in the package should be provided. For popular datasets, \url{paperswithcode.com/datasets} has curated licenses for some datasets. Their licensing guide can help determine the license of a dataset.
        \item For existing datasets that are re-packaged, both the original license and the license of the derived asset (if it has changed) should be provided.
        \item If this information is not available online, the authors are encouraged to reach out to the asset's creators.
    \end{itemize}

\item {\bf New assets}
    \item[] Question: Are new assets introduced in the paper well documented and is the documentation provided alongside the assets?
    \item[] Answer: \answerNA{}
    \item[] Justification: No new datasets or pre-trained models are released in this submission. The trained LoRA adapters and code will be released upon acceptance with accompanying documentation.
    \item[] Guidelines:
    \begin{itemize}
        \item The answer \answerNA{} means that the paper does not release new assets.
        \item Researchers should communicate the details of the dataset\slash code\slash model as part of their submissions via structured templates. This includes details about training, license, limitations, etc. 
        \item The paper should discuss whether and how consent was obtained from people whose asset is used.
        \item At submission time, remember to anonymize your assets (if applicable). You can either create an anonymized URL or include an anonymized zip file.
    \end{itemize}

\item {\bf Crowdsourcing and research with human subjects}
    \item[] Question: For crowdsourcing experiments and research with human subjects, does the paper include the full text of instructions given to participants and screenshots, if applicable, as well as details about compensation (if any)?
    \item[] Answer: \answerNA{}
    \item[] Justification: This paper does not involve crowdsourcing or research with human subjects.
    \item[] Guidelines:
    \begin{itemize}
        \item The answer \answerNA{} means that the paper does not involve crowdsourcing nor research with human subjects.
        \item Including this information in the supplemental material is fine, but if the main contribution of the paper involves human subjects, then as much detail as possible should be included in the main paper. 
        \item According to the NeurIPS Code of Ethics, workers involved in data collection, curation, or other labor should be paid at least the minimum wage in the country of the data collector. 
    \end{itemize}

\item {\bf Institutional review board (IRB) approvals or equivalent for research with human subjects}
    \item[] Question: Does the paper describe potential risks incurred by study participants, whether such risks were disclosed to the subjects, and whether Institutional Review Board (IRB) approvals (or an equivalent approval/review based on the requirements of your country or institution) were obtained?
    \item[] Answer: \answerNA{} 
    \item[] Justification: The paper does not involve crowdsourcing nor research with human subjects.
    \item[] Guidelines:
    \begin{itemize}
        \item The answer \answerNA{} means that the paper does not involve crowdsourcing nor research with human subjects.
        \item Depending on the country in which research is conducted, IRB approval (or equivalent) may be required for any human subjects research. If you obtained IRB approval, you should clearly state this in the paper. 
        \item We recognize that the procedures for this may vary significantly between institutions and locations, and we expect authors to adhere to the NeurIPS Code of Ethics and the guidelines for their institution. 
        \item For initial submissions, do not include any information that would break anonymity (if applicable), such as the institution conducting the review.
    \end{itemize}

\item {\bf Declaration of LLM usage}
    \item[] Question: Does the paper describe the usage of LLMs if it is an important, original, or non-standard component of the core methods in this research? Note that if the LLM is used only for writing, editing, or formatting purposes and does \emph{not} impact the core methodology, scientific rigor, or originality of the research, declaration is not required.
    \item[] Answer: \answerNA{} 
    \item[] Justification: The LLM is used only for writing, editing, or formatting purposes and does \emph{not} impact the core methodology, scientific rigor, or originality of the research.
    \item[] Guidelines:
    \begin{itemize}
        \item The answer \answerNA{} means that the core method development in this research does not involve LLMs as any important, original, or non-standard components.
        \item Please refer to our LLM policy in the NeurIPS handbook for what should or should not be described.
    \end{itemize}

\end{enumerate}

\end{document}